%% file: main.tex
\documentclass[journal]{IEEEtran} 

\IEEEoverridecommandlockouts                              

\usepackage{cite}
\usepackage{algorithmic}
\usepackage{graphicx}
\usepackage{mathrsfs}
\usepackage{textcomp}
\usepackage{xcolor}
\usepackage{comment}
\usepackage{amssymb}

\usepackage{amsmath, amsthm}

\usepackage{siunitx}
\usepackage{soul}
\usepackage{multirow}
\usepackage{subcaption}
\usepackage{siunitx}
\usepackage[font=footnotesize]{caption}
\usepackage[colorlinks=true,linkcolor=black,anchorcolor=black,citecolor=black,filecolor=black,menucolor=black,runcolor=black,urlcolor=black]{hyperref}
\usepackage{mathtools}
\usepackage[ruled,vlined,linesnumbered]{algorithm2e}
\usepackage{bm}
\usepackage[normalem]{ulem}
\usepackage{wrapfig}

\usepackage{nomencl}
\usepackage{etoolbox}

\makeatletter
\newenvironment{ntp}{%
  \@beginparpenalty\@lowpenalty
  \bfseries\small\textit{Note to Practitioners}\textemdash
  \@endparpenalty\@M}%
{\par}
\makeatother
\input{macros}

\title{\LARGE \bf
Socially Acceptable Bipedal Robot Navigation via Social Zonotope Network Model Predictive Control
}

\author{Abdulaziz Shamsah$^{1,2}$,
Krishanu Agarwal$^{3}$, Nigam Katta$^{3}$, Abirath Raju$^{4}$, Shreyas Kousik$^{1,*}$, and Ye Zhao$^{1,*}$
\thanks{$^{1}$George W. Woodruff School of Mechanical Engineering, Georgia Institute of Technology, Atlanta, GA, 30332-0405 USA. \texttt{ashamsah3@gatech.edu}}
\thanks{$^{2}$Mechanical Engineering Department, College of Engineering and Petroleum, Kuwait University, PO Box 5969, Safat, 13060, Kuwait}
\thanks{$^{3}$School of Electrical and Computer
Engineering, Georgia
Institute of Technology, Atlanta, GA 30308, USA.}
\thanks{$^{5}$Wallace H. Coulter Department of Biomedical Engineering, Georgia
Institute of Technology, Atlanta, GA 30308, USA.}
\thanks{$^{*}$Co-advising authorships.}
}

\begin{document}

\maketitle


\input{sections/00_abstract}

\input{sections/01_ntp}


\input{sections/02_intro}
\input{sections/03_related_work}

\input{sections/04_problem_formulation}

\input{sections/05_preliminaries}
\input{sections/06_learning_arch}
\input{sections/07_zonotope_refinement}
\input{sections/08_MPC}

\input{sections/09_implementation}

\input{sections/10_results}

\input{sections/11_hardware}
\input{sections/12_limitations}

\input{sections/13_conclusion}

\appendices
\input{appendix}
\input{Acknowledgment}



\bibliographystyle{IEEEtran}
\bibliography{lidar.bib}

\end{document}

%% file: macros.tex
\newtheorem{thm}{Theorem}[section]

\newtheorem{prop}[thm]{Proposition}
\newtheorem{prob}[thm]{Problem}

\newtheorem{assum}[thm]{Assumption}

\newtheorem{defn}{Definition}[section]

\newtheorem{rem}{Remark}

\providecommand{\N}{\ensuremath \mathbb{N}}

\newcommand{\set}[1]{\mathcal{#1}}

\newcommand{\state}{\boldsymbol{x}}

\newcommand{\ctrl}{\boldsymbol{u}}

\newcommand{\position}{\boldsymbol{p}}

\newcommand{\ego}{^{\rm ego}}
\newcommand{\ped}{^{p_k}}
\newcommand{\future}{_{[t, t_f]}}
\newcommand{\past}{_{[t_p, t]}}
\newcommand{\loss}{\mathcal{L}}

\newcommand{\pastq}{_{[t_p, t],q}}

\newcommand{\nextq}{_{q+1}}
\newcommand{\currq}{_{q}}

\newcommand{\zonotope}{\set{Z}}
\newcommand{\setfn}[1]{\mathscr{#1}}
\newcommand{\zonofnplain}{{\setfn{Z}\!}}
\newcommand{\zonofn}[1]{{\zonofnplain\!\left(#1\right)}}
\newcommand{\centers}{\boldsymbol{c}}
\newcommand{\mink}{^{\rm mink}}

\newcommand{\local}{^{\rm loc}}

\newcommand\Tstrut{\rule{0pt}{2.6ex}}         
\newcommand\Bstrut{\rule[-0.9ex]{0pt}{0pt}}

%% file: sections/00_abstract.tex
\begin{abstract}
This study addresses the challenge of social bipedal navigation in a dynamic, human-crowded environment, a research area 
largely underexplored in legged robot navigation.
We present a zonotope-based 
framework that 
couples prediction and motion planning for a bipedal ego-agent to 
account for bidirectional influence with the surrounding pedestrians.
This framework incorporates a Social Zonotope Network (SZN), a neural network that 
predicts future pedestrian reachable sets 
and plans future socially acceptable reachable set for the ego-agent.
SZN generates the reachable sets as zonotopes for efficient reachability-based planning, collision checking, and online uncertainty parameterization.
Locomotion-specific losses are added to the SZN training process to adhere to  
the dynamic limits of the bipedal robot that are not explicitly present in the human crowds data set. These loss functions enable the SZN to generate locomotion paths that are more dynamically feasible for improved tracking.
SZN is 
integrated with a Model Predictive Controller (SZN-MPC) for footstep planning for our bipedal robot Digit. 
SZN-MPC solves for collision-free 
trajectory by optimizing through 
SZN's gradients. and Our results demonstrate the framework's effectiveness in producing a socially acceptable path, with consistent locomotion velocity,  and optimality. 
The SZN-MPC framework is validated with extensive simulations and hardware experiments. 
\end{abstract}

%% file: sections/01_ntp.tex
\vspace{0.1in}
\begin{ntp}
This paper is motivated by the challenge of navigating bipedal robots through dynamic, human-crowded environments in a socially acceptable manner.
Existing methods for social navigation often only address obstacle avoidance and are demonstrated on a robot with simple dynamics.
This paper proposes the Social Zonotope Network (SZN), a novel neural network that couples pedestrian future trajectory prediction and robot motion planning to facilitate socially aware navigation for bipedal robots such as Digit, designed by Agility Robotics.

The social behaviors are learned from real open-sourced pedestrian data using the SZN, which outputs the future predictions as reachable sets for each agent in the environment.
The SZN is then integrated into a trajectory optimization problem that takes into account personal space preferences and bipedal robot capabilities to design trajectories that are both collision-free and socially acceptable. This work also highlights the computational efficiency of the SZN design that makes it suitable for real-time integration with motion planners.
The framework is validated through extensive simulations and hardware experiments.

From a practical standpoint, this research provides a framework that can be applied to bipedal robots to improve automation in human-populated environments such as hospitals, shopping centers, and airports.
The framework's ability to automatically adapt to surrounding human movement helps minimize disruptions and ensures that the robot's presence is not a hindrance to the flow of human traffic.
Future work will focus on outdoor deployment, which will require onboard perception capabilities to detect surrounding pedestrians. 

\end{ntp}

%% file: sections/02_intro.tex
\begin{figure}[t]
\centerline{\includegraphics[width=.75\columnwidth]{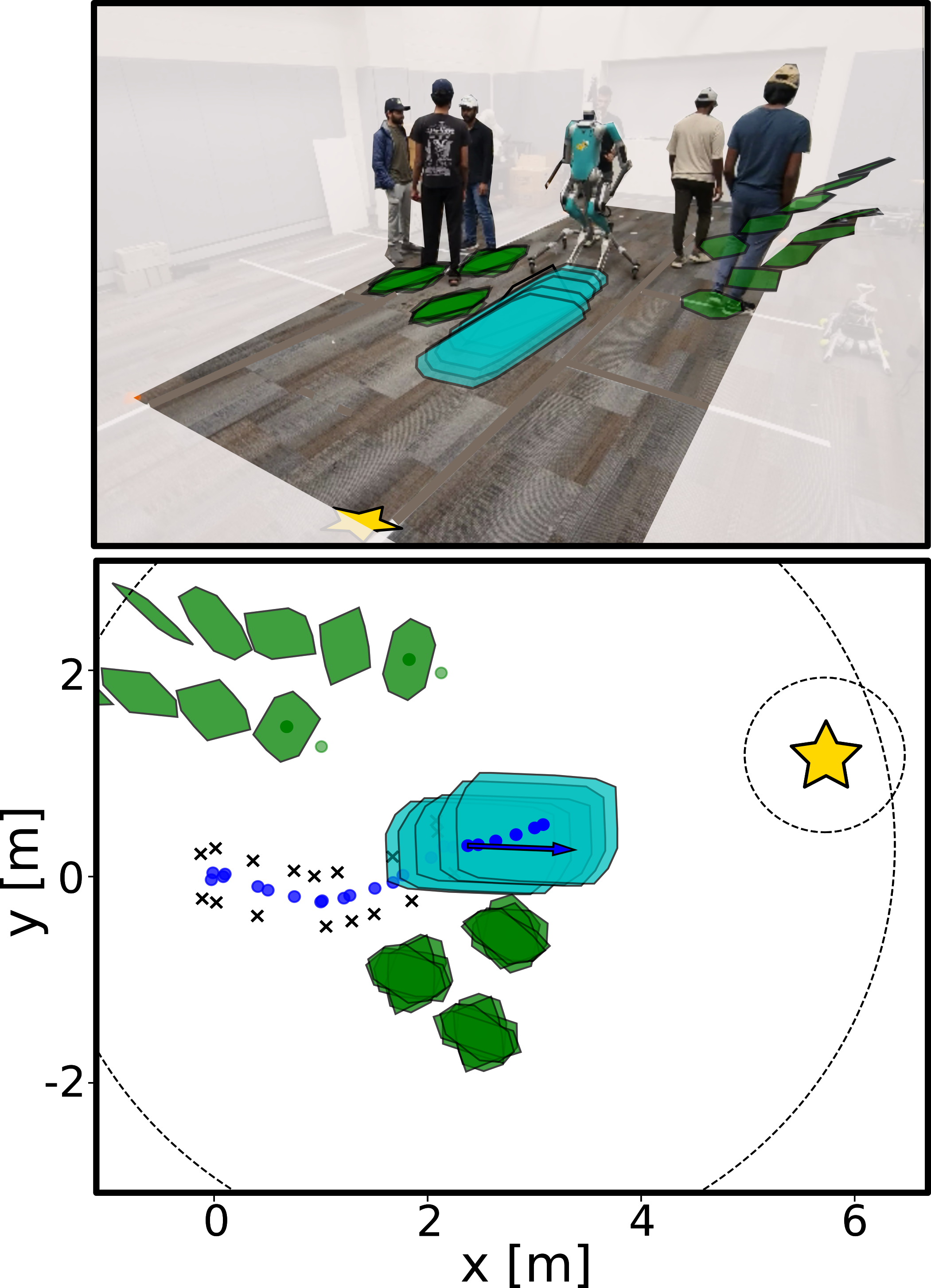}}
\caption{(top) Snapshot of the proposed social path planner demonstrated on hardware with $5$ pedestrians with pedestrian's prediction zonotopes (green), ego-agent's social zonotope (cyan), and goal location (yellow star) superimposed. (bottom) shows a top-down view of the ego-agent's path, pedestrians' prediction, and social zonotopes.}
\label{fig:high-level}
\end{figure}

\begin{figure*}[t]
\centerline{\includegraphics[width=.9\textwidth]{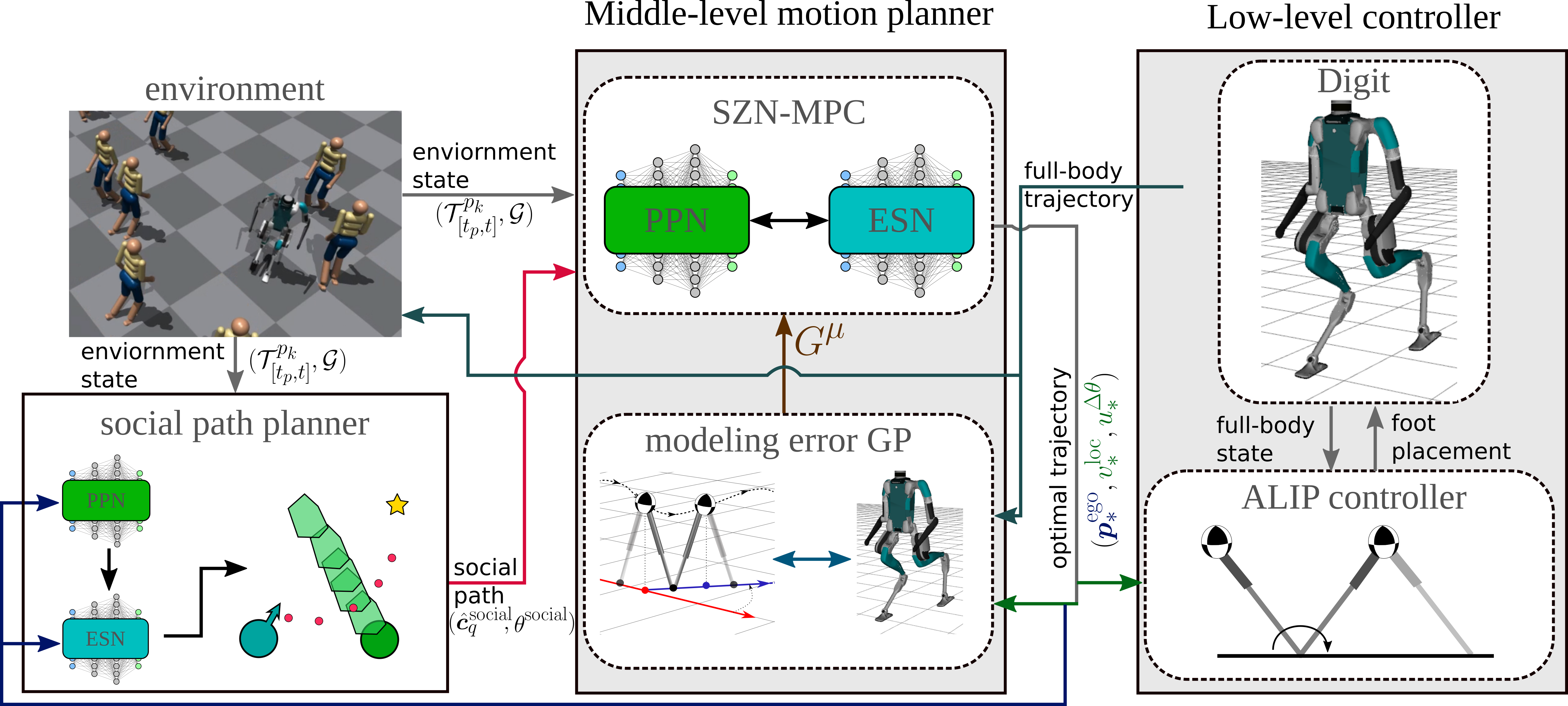}}
\caption{Block diagram of the proposed framework. The framework is developed around the Social Zonotope Network (SZN) (Sec.~\ref{sec:social_zono_net}), which is composed of two sub-networks: the Pedestrian Prediction Network (PPN) and the Ego-agent Social Network (ESN) shown in green and cyan, respectively. 
Given an environment with observed pedestrians and a goal location, the social path planner designs a social path for Digit (Sec.~\ref{subsec:cost}).
At the middle level, SZN-MPC optimizes through SZN to generate both collision-free and socially acceptable trajectories for Digit (Sec.~\ref{sec:SDMPC}).
The optimal trajectory is then sent to the ALIP controller~\cite{Gong2022AngularMomentum} to generate the desired foot placement for reduced-order optimal trajectory tracking.
An ankle-actuated-passivity-based controller~\cite{sadeghian2017passivity,shamsah2023integrated} is implemented on Digit for full-body trajectory tracking.
Digit current velocity and the optimal trajectory from SZN-MPC are used in the modeling error GP to compensate for the modeling uncertainty between ROM dynamics and full-order dynamics (Sec.~\ref{subsec:Modelling_error_refinements}).}
\label{fig:block_framework}
\end{figure*}

\section{Introduction}
Bipedal locomotion in complex environments has garnered substantial attention in the robotics community~\cite{li2023autonomous, huang2023efficient, narkhede2022sequential, gibson2022terrain, zhao2022reactive, Kulgod2020LTL, warnke2020towards, zhao2017robust}.
Social navigation poses a particular challenge due to the inherent uncertainty of the environment, unknown pedestrian dynamics, and implicit social behaviours~\cite{mavrogiannis2023core}. 
Recently, there has been an increasing focus on social navigation for mobile robots in human environments~\cite{che2020efficient, bera2019emotionally, paez2022unfreezing, nishimura2020risk, schaefer2021leveraging, majd2021safe, cathcart2023proactive}. Using mobile robots benefits from stable motion dynamics, which facilitate the investigation of high-level social path planners.
In contrast, the exploration of social navigation for bipedal robots remains unexplored.
This is largely attributed to the intricacies of the hybrid, nonlinear, and high degrees-of-freedom dynamics associated with bipedal locomotion.
The complexities inherent in stabilizing and controlling bipedal robots have led researchers to prioritize fundamental locomotion research in the past~\cite{gu2023walking, gu2022reactive,yang2022bayesian,DataS2S_Dai}, delaying advancements in deploying these robots into more complex dynamic environments such as human-populated ones (see Fig.~\ref{fig:high-level}).

In this study, we present an integrated framework for prediction and motion planning for socially acceptable bipedal navigation as shown in Fig.~\ref{fig:block_framework}.
We propose a Social Zonotope Network (SZN) that both predicts pedestrians' future trajectories and plans a socially acceptable path for the bipedal robot Digit~\cite{agility} with 28 degrees of freedom (DoFs).
SZN is trained with locomotion-specific losses to improve full-body tracking of the desired socially acceptable paths.

Our SZN model outputs reachable sets for both pedestrians and the ego-agent represented as zonotopes, a class of convex symmetric polytopes.
Zonotopes enable efficient and robust reachability-based planning, collision checking, and uncertainty parameterization~\cite{paparussozapp, selim2022safe, kousik2019safe, althoff2010reachability}.
In this work, we detect and avoid collisions between zonotopes corresponding to the ego-agent and pedestrians.
Additionally, the ego-agent's zonotope is refined to compensate for (1) robot modeling errors of the ego-agent's dynamics via Gaussian process (GP) regression, and (2) ego-agent's personal space for increased social acceptance in practice.
Zonotopes provide a balance between geometric complexity and computational efficiency.
We specifically take advantage of two facts: (1) the Minkowski sum of two zonotopes is again a zonotope, allowing us to easily augment the zonotopes output by a neural network;  and (2) collision checking a pair of zonotopes can be differentiated for use in gradient-based motion planning methods.

Our framework integrates SZN in a model predictive controller (MPC) at the middle level as shown in Fig.~\ref{fig:block_framework}.
SZN-MPC optimizes over the output of the neural network,  by encoding it as reachability and collision avoidance constraints. We incorporate a novel cost function designed to encourage the generation of socially acceptable trajectories.
SZN-MPC employs a reduced-order model (ROM), i.e., a Linear Inverted Pendulum (LIP) model, for the bipedal locomotion process, and then sends optimal commands, i.e., center of mass (CoM) velocity and heading change, to the low-level controller on Digit for full-body joint trajectory design and control.

\subsection{Contributions and Outline}
The main contributions of this study are as follows.
\begin{itemize}
\item A reachability-based prediction and planning framework for bipedal navigation in a social environment: we introduce the Social Zonotope Network (SZN), a CVAE architecture for coupled pedestrian future trajectory prediction and ego-agent social planning both parameterized as zonotopes.

\item Learning locomotion safety: we encode locomotion safety into the learning module using signal temporal logic (STL) specifications, and design its robustness score as loss functions of the CVAE network during training.

\item Online zonotope refinements for social acceptability and model discrepancy compensation: we leverage the adaptable
nature of zonotope parameterization to enhance social acceptability by incorporating personal space refinements and to efficiently adjust online for learned model discrepancies between the reduced-order model (ROM) and the full-order model of Digit.
\item SZN-MPC: we integrate the SZN with a MPC, where the zonotopes outputted by SZN are encoded as constraints for reachibility-based motion planning and collision checking.
\item Social acceptability cost function design: a novel MPC cost function to plan trajectories for the ego-agent using learned paths from human data sets.

\item Experimental hardware evaluation: we implement the proposed framework on a bipedal robot Digit equipped with a low-level passivity controller for full-body joint control.

\end{itemize}

This article is organized as follows.
Sec.~\ref{sec:related_work} is a literature review of related work.
Sec,~\ref{sec:prob_formulation} introduces the problem we are seeking to solve.
Then, the robot dynamics, environment setup and zonotope preliminaries are in Sec.~\ref{sec:prelim}.
Sec.~\ref{sec:social_zono_net} presents SZN's architecture and loss functions. Zonotope refinements for social acceptability and uncertainty parameterization in Sec.~\ref{sec:zonotope_refinement}. 
Sec.~\ref{sec:SDMPC} formulates the problem as a MPC and introduces the social acceptability cost.
Implementation details and results are in Sec.~\ref{sec:results}.
In Sec.~\ref{sec:discussion} we discuss the limitations of the proposed framework.
Finally, Sec.~\ref{sec:conclusion} concludes this article.

This paper expands upon a previous conference version~\cite{shamsah2024real}.
The work presented here extends the SZN training to include additional locomotion-specific loss functions, incorporates a social acceptability cost function in the SZN-MPC, and introduces a coupled SZN-MPC for simultaneous pedestrian prediction and ego-agent planning.
Additionally, this paper refines the zonotope parameterization by incorporating personal space modulation and uncertainty parameterization. We also benchmark the collision-avoidance performance of our method with a control barrier function baseline and validate our framework by extensive hardware experiments that were not included in the conference version.

%% file: sections/03_related_work.tex
\section{Related Work}\label{sec:related_work}

This work lies in the intersection of social navigation, human trajectory prediction, gradient-based safe motion planning, and bipedal locomotion.
We now review each of these topics.

\begin{figure}[t]
\centerline{\includegraphics[width=.99\columnwidth]{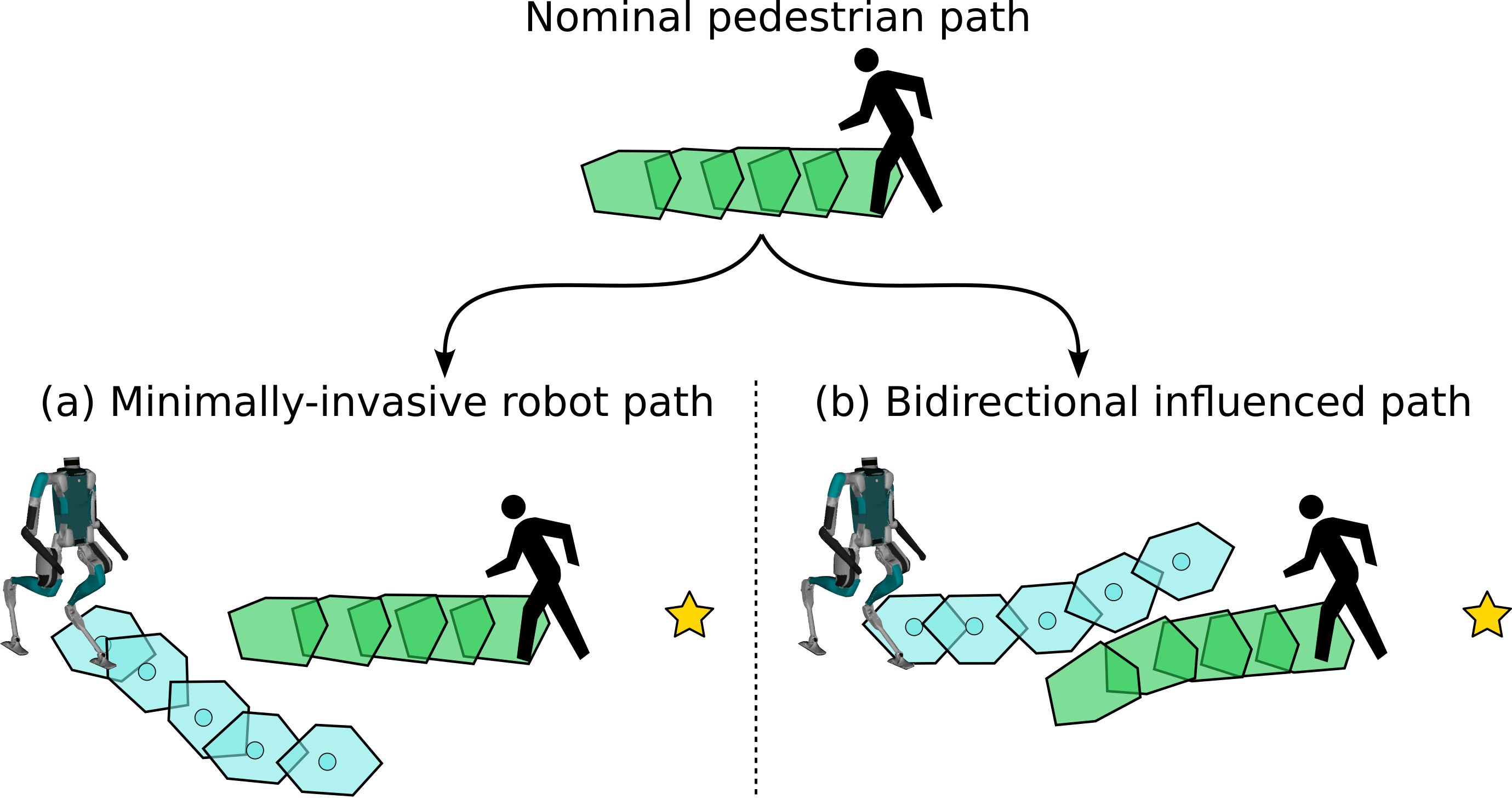}}
\caption{Comparison between two different designs of socially acceptable paths. (a) shows a minimally invasive path design for the robot as in~\cite{moder2022proactive, schaefer2021leveraging}, i.e., the robot adjusts its path to not alter  the pedestrian's original path or change it in a minimally-invasive way. (b) shows the bidirectionally influenced path between the robot and the pedestrian that our work employs, i.e., the robot and the pedestrian mutually react to each other and adjust their own paths accordingly. }
\label{fig:bidirectional}
\end{figure}

\subsection{Social Navigation}\label{RW:social_nav}

Social navigation literature can be categorized into two main categories coupled prediction and planning and decoupled prediction and planning~\cite{mavrogiannis2023core}. In this section, we will present different approaches in the literature and conclude where our proposed work falls in the social navigation field.
\subsubsection{Coupled Prediction and Planning}
In coupled prediction and planning literature, the proposed frameworks recognize the influence of the robot motion on the surrounding pedestrians and vice versa~\cite{mavrogiannis2023core}.
Navigating an environment with humans in a socially compliant manner requires a proactive approach to motion planning \cite{truong2017social,moder2022proactive, schaefer2021leveraging, cathcart2023proactive}.
In~\cite{cathcart2023proactive}, the authors use coupled opinion dynamics to proactively design motion plans for a mobile robot, without the need for human prediction models.
It's an implicit approach, as it relies only on the observation of the approaching human position and orientation to form an opinion that alters the nominal path and avoids collisions with pedestrians.
Another implicit approach is~\cite{kretzschmar2016socially}, which uses inverse reinforcement learning to learn robot motions that mimics human behavior.

On the other hand, the approaches in~\cite{moder2022proactive, schaefer2021leveraging} are explicit, as the framework explicitly predicts the future trajectories of pedestrians.
The authors in~\cite{moder2022proactive} propose a social interference metric based on Kullback-Leibler divergence to measure the interference of the robot's path plan on the surrounding humans' future trajectory.
Hypothesizing that minimizing the social interference metric will result in a socially acceptable trajectory for the ego-agent.
Similarly, a gradient-based trajectory optimization is introduced in \cite{schaefer2021leveraging} to minimize the difference between the humans' future path prediction conditioned on the robot's plan and the nominal prediction.

The studies of~\cite{moder2022proactive, schaefer2021leveraging} both work under the assumption that a minimally-invasive robot trajectory, with minimal effect on surrounding humans' nominal trajectory, is socially acceptable (see Fig.~\ref{fig:bidirectional}(a)).
In contrast, our work leverages a different notion of social acceptability, where we aim to learn the socially acceptable trajectory of the ego-agent from human crowd datasets to avoid any heuristics or biases on prioritizing the minimal invasiveness of the pedestrians as the socially acceptable metric. Our method enables bidirectional influence between the pedestrians and the ego-agent, and allows the ego-agent to change surrounding pedestrians' paths (see Fig.~\ref{fig:bidirectional}(b)).

\subsubsection{Decoupled Prediction and Planning}
In decoupled prediction and planning, social navigation can be viewed as dynamic obstacle avoidance~\cite{mavrogiannis2023core}. 
The work in~\cite{brito2019model} ignores any effect the robot path has on the surrounding pedestrians. They propose a simple linear Kalman filter to predict the pedestrians' future paths and an MPC for collision avoidance. This approach while effective in a small number of pedestrians, ignoring the coupling effect between the robot and the pedestrians will result in the reciprocal dance problem, i.e., an oscillatory interaction between the robot and the pedestrians~\cite{feurtey2000simulating}, or the so-called "freezing robot problem"~\cite{trautman2015robot}, i.e., the robot comes to a stop to avoid collisions.

Based on the aforementioned categories, our framework uses a coupled prediction and planning approach, as the pedestrians' future predictions are conditioned on the ego-agent's future planned motion.
Our approach integrates both explicit and implicit elements. Explicitly, we predict the future trajectories of pedestrians and utilize these predictions for collision avoidance.
Implicitly, we leverage learned trajectories from pedestrian datasets to generate ego-agent trajectories that mimic human behavior.

\subsection{Human Trajectory Prediction}
Our framework is inspired by the human trajectory prediction community, such as Trajectron++~\cite{salzmann2020trajectron++}, SocialGAN~\cite{gupta2018social}, PECNet~\cite{mangalam2020not}, Y-net~\cite{mangalam2021goals}, Sophie~\cite{sadeghian2019sophie}, and STAR~\cite{yu2020spatio} where we aim to design a socially acceptable trajectory for the ego-agent that mimics the path from human crowd datasets.
The work in \cite{hong2023obstacle} proposes an obstacle avoidance learning method that uses a Conditional Variational Autoencoder (CVAE) framework to learn a temporary, near-horizon target distribution to avoid pedestrians actively.
However, during the training phase, the temporary targets are selected heuristically.
In contrast, we aim to learn such temporary waypoints from human crowd datasets to capture a heuristic-free socially acceptable path.
In~\cite{mangalam2020not}, the authors develop a simple but accurate CVAE architecture based on Multi-Layer Perceptrons (MLP) networks to predict crowd trajectories conditioned on past observations and intermediate endpoints. Our SZN inherits a similar MLP-based CVAE architecture, where the ego-agent path is conditioned on the final goal location and surrounding pedestrians' future trajectories. In addition, the SZN is also conditioned on the immediate change in the ego-agent state to be better integrated with MPC for planning.
Leveraging a simple network architecture in our work, i.e., MLP, instead of other advanced architecture such as LSTMs~\cite{salzmann2020trajectron++,paparussozapp} or Transformers~\cite{yu2020spatio}, enables real-time planning and prediction when integrated into gradient-based motion planning for the ego-agent.

\subsection{Neural-Network-based Motion Planning}

The work in~\cite{schaefer2021leveraging, paparussozapp} both leverage the neural network gradients into a trajectory optimization (TO) problem for safe motion planning.
The method introduced in~\cite{schaefer2021leveraging} employs~\cite{salzmann2020trajectron++} for generating multimodal probabilistic predictions and integrates the prediction model gradients in the cost function of the TO problem, to minimize the invasiveness of the robot's path to the surrounding pedestrians' paths (as previously discussed in Sec.~\ref{RW:social_nav}).

However, most prediction frameworks generate a discrete time representation of the trajectories posing challenges to guarantee safety.
Leveraging zonotopes as reachable sets provides a robust framework for ensuring safety in robot models by effectively capturing inherent uncertainties in both models and dynamics \cite{selim2022safe, kousik2019safe, paparussozapp}.
The work of \cite{paparussozapp} presents a Zonotope Alignment of Prediction and Planning (ZAPP) that relies on zonotopes to enable continuous-time reasoning for planning.
They use Trajectron++ \cite{salzmann2020trajectron++} to predict the obstacle trajectories as a Gaussian distribution.
They construct a zonotope over these distributions, which leads to an overapproximation of the uncertainties. The predictive model gradients are used in the TO constraints for obstacle avoidance~\cite{paparussozapp}, rather than in the cost function as designed in \cite{schaefer2021leveraging}.

Compared to~\cite{schaefer2021leveraging,paparussozapp}, we propose learning path prediction distributions directly as zonotopes, bypassing the initial step of predicting Gaussian distributions for pedestrian motion.
As such, our approach is computationally efficient and facilitates real-time integration with an MPC, while an online implementation of \cite{paparussozapp} is challenging. The learned zonotopes also provide the gradients for the constraints in the proposed MPC for reachability-based planning and collision checking.

\subsection{Bipedal Locomotion}
Bipedal locomotion has been extensively studied with a wide spectrum of model representations and methods---such as reduced-order model (ROM)~\cite{huang2023efficient,narkhede2022sequential,zaytsev2018boundaries, gibson2022terrain}, single rigid body model \cite{Ding_MPC_SRBM}, centroidal model \cite{Centroidal_MPC}, whole-body models~\cite{yang2022bayesian, DataS2S_Dai, smit2019walking}, ROM-inspired reinforcement learning \cite{castillo2023template}, and model-free reinforcement learning approaches~\cite{li2021reinforcement, radosavovic2023learning}---to name a few.
Generally, locomotion planning using whole-body models renders a high computation cost~\cite{hereid2018dynamic, grizzle2014models} and becomes much less efficient for navigation in complex environments that often involve long-horizon planning; therefore, in this work, we focus on ROM-based methods for CoM trajectory and foot placement planning.

The work in\cite{huang2023efficient} uses an omnidirectional differential-drive wheeled robot model with a preference for sagittal movement to capture bipedal robots' behaviors accurately. 
They design a control Lyapunov function (CLF) to drive and orient that robot towards the goal and a control barrier function (CBF) for obstacle avoidance.
The framework is demonstrated for indoor and outdoor navigation in environments with static obstacles only. 
In our work, we use Linear Inverted Pendulum (LIP)~\cite{kajita20013d} as ROM for our bipedal robot.
A similar model is used  in~\cite{narkhede2022sequential}, while~\cite{teng2021toward, gibson2022terrain} use an angular momentum-based LIP (ALIP)~\cite{Gong2022AngularMomentum} for improved prediction of CoM velocities.

The authors in~\cite{narkhede2022sequential} present discrete-time CBF (DCBF)~\cite{agrawal2017discrete} constraints with a sequential LIP-based MPC to plan trajectories for Digit in an environment with static and dynamic obstacles.
However, unlike our approach where we predict the future trajectory of the pedestrians, the dynamic obstacles trajectory is assumed to be known in~\cite{narkhede2022sequential} and the framework is only demonstrated in simulation.
The work in~\cite{teng2021toward} also uses a DCBF-MPC to avoid static obstacles in simulation.

In contrast to the aforementioned studies, the overarching goal of this study is to develop a bipedal locomotion planner that can generate safe paths while considering social norms and can be efficiently solved in real-time for hardware implementation.
To this end, we propose SZN-MPC, a LIP-based MPC with SZN to predict reachable sets of the obstacles for collision avoidance and to plan a socially acceptable, zonotope-based path for the ego-agent.
SZN-MPC computes the desired ego-agent trajectory, which is then sent to the ALIP controller~\cite{Gong2022AngularMomentum} for foot placement generation and the passivity controller~\cite{sadeghian2017passivity} for low-level joint control with ankle-actuation for improved ROM tracking~\cite{shamsah2023integrated}.

%% file: sections/04_problem_formulation.tex
\section{Problem formulation}
\label{sec:prob_formulation}

We seek to safely navigate a bipedal robot through a crowd.
We now present our robot model and environment model, and then formalize this problem.
\subsection{Robot Model}
\begin{figure}[t]
\centerline{\includegraphics[width=.9\columnwidth]{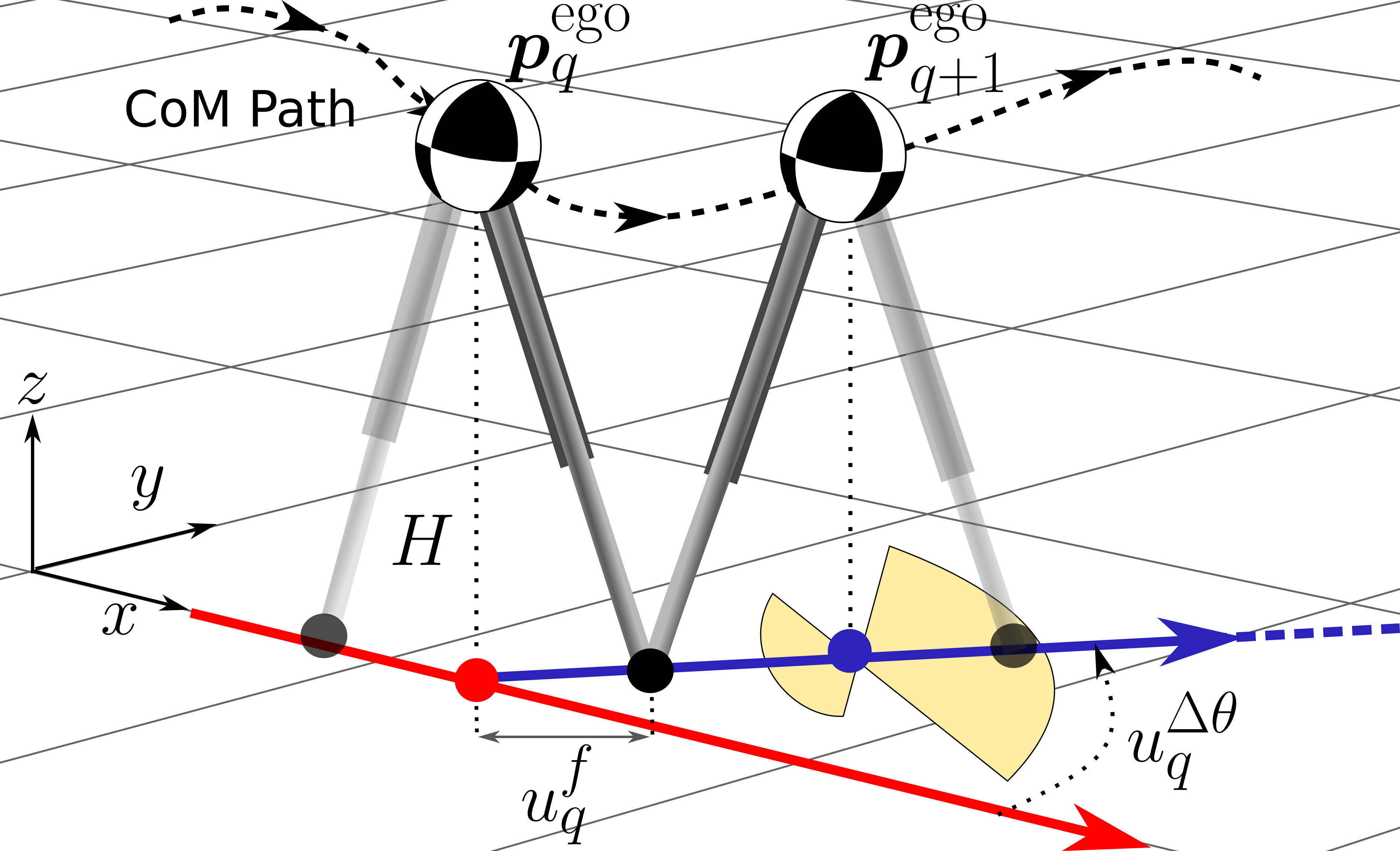}}
\caption{Illustration of the Linear Inverted Pendulum model for two consecutive foot contact switching states $\state \currq$ and $\state \nextq$.
The shaded yellow regions indicate the kinematics constraint on the control input $\ctrl$ detailed in Sec.~\ref{subsec:kin_const}.}
\label{fig:LIP}
\end{figure}
We leverage a locomotion-specific reduced order model (ROM) to describe our bipedal robot dynamics in this study. Consider a bipedal ego-agent with discrete time dynamics  $\state\nextq = \Phi(\state\currq, \ctrl \currq)$,
where $\state\currq$ and $\ctrl \currq$ are the state and control input respectively at the $q^{\rm th}$ walking step\footnote{The robot model used in our study represent step-by-step dynamics, i.e., $\state\currq$ and $\state\nextq$ are the CoM state at the foot contact switching instant of two consecutive walking steps}.
The state of the bipedal robot, i.e., the ego-agent, is $\state=(\position \ego, v\local, \theta)$ where $\position \ego = (x,y)$ is the 2-D location in the world coordinate, $v\local$ is the sagittal velocity at the foot contact switching instant in the local coordinate, and $\theta$ is the heading.
The control input is $\ctrl \currq = (u^{f}_{q} \quad u^{\Delta \theta} \currq)$, where $u^{f}_{q}$ is the sagittal foot position relative to the CoM, and $u^{\Delta \theta} \currq$ is the heading angle change between two consecutive walking steps. A schematic robot model is  shown in Fig.~\ref{fig:LIP}.

\subsection{Environment Setup and Problem Statement}
The ego-agent 
is tasked to navigate to a known goal location $\set{G}$ in an open environment with $m \in \N$ observed pedestrians treated as dynamic obstacles.
Denote the pedestrian state $\set{T}\ped\past $, which is the 2-D trajectory of $k^{\rm th}$ pedestrian observed over a discrete time interval $[t_p, t]$, where $t_p < t$.
The environment is partially observable as only the pedestrians in a pre-specified sensory radius of the ego-agent are observed.
The path the ego-agent takes, should ensure locomotion safety, navigation safety, as well as promoting social acceptability.
\begin{defn}[Locomotion safety] For bipedal robots, locomotion safety is defined as maintaining balance dynamically throughout its locomotion process. 
\end{defn}

\begin{defn}[Navigation safety] Navigation safety is defined as maneuvering in human crowded environments while avoiding collisions with pedestrians, i.e., $\|\position\ego_t -  \set{T}\ped_t \| > d, \; \forall\ t, k \in m$, where $\position\ego_t$ denotes the ego-agent 2-D position and $d$ represent the minimum allowable distance between the ego-agent and the pedestrians.
\label{def:nav_safety}
\end{defn}

\begin{defn}[Socially acceptable path for bipedal systems] A path that a bipedal ego-agent takes in a human-crowded environment is deemed socially acceptable if it has an Average Displacement Error (ADE) $ < \epsilon$ when compared to ground truth human data navigating in the same environment\footnote{$\epsilon$ represents the allowable deviation from the socially acceptable path. The Average Displacement Error denotes the average error between the planned path and the ground-truth path.}. 
\label{def:social_path}
\end{defn}

Based on the aforementioned definitions and environment setup the problem we seek to solve is as follows:
\begin{prob}
Given the discrete dynamics of the bipedal robot $\state \nextq = \Phi(\state \currq, \ctrl \currq)$ and an environment state $\set{E}=(\set{T}\ped\past,\set{G})$, find an optimal motion plan for the bipedal robot and a navigation path that promotes social acceptability for the bipedal ego-agent in a partially observable environment containing pedestrians while ensuring locomotion and navigation safety. 
\label{problem_statment}
\end{prob}

%% file: sections/05_preliminaries.tex
\section{Preliminaries}
\label{sec:prelim}
This section introduces the dynamic model used for the robot, learning and environment assumptions, and zonotopes.
\subsection{Robot Walking Model}
\label{subsec:ROM}
The ROM used to design the 3-D walking motion of Digit, as introduced in the problem formulation (Sec.~\ref{sec:prob_formulation}), is the linear inverted pendulum (LIP) model \cite{kajita20013d}. For the LIP model we assume that each walking step has a fixed duration\footnote{set to be equal to the timestep between frames in the dataset ($0.4$ s)} $T$~\cite{narkhede2022sequential, teng2021toward}.
Then we build our model on the discrete sagittal dynamics\footnote{the lateral dynamics are only considered in the ALIP model at the low level since they are periodic with a constant desired lateral foot placement (See Fig.~\ref{fig:block_framework})} $(\Delta x\local \currq,v\local\currq)$, where $x\local\currq$ is CoM position at the beginning of the $q^{\rm th}$ step, $\Delta x\local=x\local\nextq - x\local\currq$ is the local sagittal CoM position difference between two consecutive walking steps, and $v\local\currq$ is the sagittal velocity at the local coordinate for the $q^{\rm th}$ walking step as shown in Fig.~\ref{fig:LIP} (see Appendix.~\ref{appendix:LIP} for detailed derivation):
\begin{equation}
    \Delta x\local (u^f\currq) = \left(v\local\currq \frac{\sinh(\omega T)}{\omega} + (1 - \cosh(\omega T))u^f\currq\right)
    \label{eq:delta_x}
\end{equation}
\begin{equation}
    v\local\nextq (u^f\currq) = \cosh(\omega T) v\local\currq - \omega \sinh(\omega T) u^f\currq 
    \label{eq:sagittal_velocity}
\end{equation}
where $\omega = \sqrt{g/H}$, $g$ is the gravitational constant, and $H$ is the constant CoM height.
The heading angle change is governed by $u^{\Delta \theta} \currq = \theta\nextq - \theta_{q}$.
Based on the sagittal dynamics~(\ref{eq:delta_x}) and~(\ref{eq:sagittal_velocity}), we introduce coordinate transformation based on the heading angle $\theta \currq$ to control the LIP dynamics in 2-D Euclidean space.
Therefore the full LIP dynamics in 2-D Euclidean space become:
\begin{subequations}
\label{eq:lip_dynamics}
\begin{align}
x\nextq &= x_q +\Delta x\local (u^f\currq)\cos(\theta_q) \\
y\nextq &= y_q + \Delta x\local (u^f\currq) \sin(\theta_q) \\
v\local\nextq  &= \cosh(\omega T) v\local\currq - \omega \sinh(\omega T) u^f\currq\\
\theta\nextq &= \theta\currq + u^{\Delta \theta}\currq 
\end{align}
\end{subequations}

A detailed derivation of~(\ref{eq:lip_dynamics}) is in Appendix.~\ref{appendix:LIP}.
For notation simplicity, and hereafter, (\ref{eq:lip_dynamics}) will be referred to as:
\begin{equation}
    \state \nextq = \Phi(\state \currq, \ctrl \currq)
    \label{eq:ROM_general}
\end{equation}
Later on,~\eqref{eq:ROM_general} will be enforced as dynamic constraints in our proposed SZN-MPC in Sec.~\ref{sec:SDMPC}.
\subsection{Environment Assumptions and Observations}
In this work, we hypothesize that in a social setting, the information accessible by a human for determining their future path is threefold: (i) their final destination $\set{G}=(x^{\rm dest}, y^{\rm dest})$ (navigation intent), (ii) the surrounding pedestrians' past trajectory\footnote{the subscripts $t_p$, $t$, and $t_f$ represent discrete time indices denoting the past, current and future trajectories, respectively, where $t_p < t < t_f$.} $\set{T} \ped \past=\{x \ped \currq, y \ped \currq\}^{t}_{q=t_p}, \forall k$, where $k$ indexes the $k^{\rm th}$ pedestrian, and (iii) the human's social experience, i.e., their assumptions on how to navigate the environment in a socially-acceptable manner.
We treat the social experience as latent information that is not readily available in human crowd datasets. Therefore we make the following assumption.
\begin{assum}

Suppose we learn a model of the future trajectory of a human as a function of their final goal $\set{G}$ and surrounding pedestrians' past trajectories $\set{T} \ped \past$. We assume that this model will implicitly represent each human's social experience.
\end{assum}

In this work, to learn a socially acceptable future path for an ego-agent $\set{T} \ego \future=\{x\ego\currq, y \ego \currq\}^{t_f}_{q=t}$, we use real human crowd datasets, and substitute a single human in the dataset as the ego-agent.

Only the pedestrians within a prespecified radius of the ego-agent are observable, 
and we assume that their past trajectories were observable over a specified time interval from $t_p$ to $t$.

\subsection{Zonotope Preliminaries}
\begin{figure}[t]
\centerline{\includegraphics[width=.8\columnwidth]{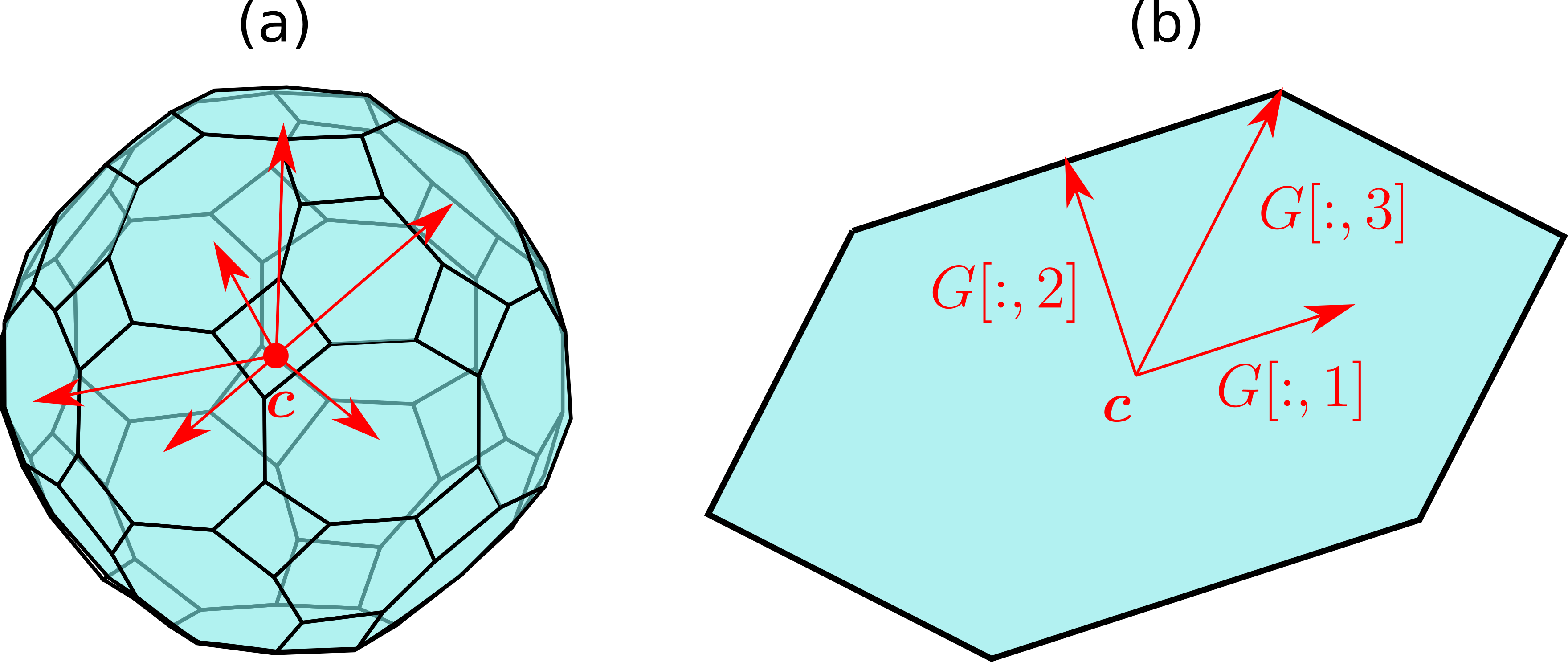}}
\caption{An illustration of zonotopes: (a) a 3-D zontope ($n=3$) with $n_G=13$ (b) a 2-D zonotope ($n=2$) with $n_G=3$. Red arrows indicate the generators in $G$, with only 6 out of 13 generators are illustrated in (a). In this study, we will use the 2-D zonotopes for our reachability path design.}
\label{fig:zonotope}
\end{figure}

A zonotope $\zonotope \in \mathbb{R}^n$ is a convex, symmetrical polytope parameterized by a center $\centers \in \mathbb{R}^n$ and a generator matrix $G \in \mathbb{R}^{n \times n_{G}}$ (see Fig.~\ref{fig:zonotope}).
\begin{equation}
    \zonotope = \zonofn{\centers, G}=\{\centers + G\beta  \; | \; \|\beta\|_\infty \leq 1 \}
\end{equation}

The Minkowski sum of $\zonotope_1 =\zonofn{\centers_1, G_1}$ and $\zonotope_2=\zonofn{\centers_2, G_2}$ is $\zonotope_1 \oplus \zonotope_2 = \mathscr{Z}{(\centers_1+\centers_2, [G_1 \; G_2])}$.
To Check collisions between two zonotopes, \cite[Lemma 5.1]{guibas2003zonotopes} is used:
\begin{prop}(\cite[Lemma 5.1]{guibas2003zonotopes})
\label{prop:intersection}
    $\zonotope_1 \cap \zonotope_1 = \emptyset $ iff $\centers_1 \notin \zonofn{\centers_2, [G_1 \; G_2]}$
\end{prop}

Per \cite[Theorem 2.1]{althoff2010reachability}, zonotopes can be parameterized using a half-space representation $\set{P} = \{x \; | \;Ax \leq b\}$, where $x \in \set{P} \iff \max (Ax - b) \leq 0$ and $ x \notin \set{P} \iff \max (Ax - b) > 0 $ (see Fig.~\ref{fig:zonotope}(b)), which we show is useful for collision checking.
In the special case of a 2-D zonotope, the center-generator representation to the half-space representation is given analytically as follows:
\begin{prop}(\cite[Proposition 2]{paparussozapp})
\label{prop:half_space}
    Let $C=\begin{bmatrix} - G[2,:]& G[1,:]]\end{bmatrix}$ and $l_{G}[i]$ be the norm of the $i^{\rm th}$ generator $l_{G}[i] = \| G[:,i] \|_2$, the half-space representation of a 2-D zonotope:
    \begin{equation}
        A[i,:] = \frac{1}{l_{G}[i]} \cdot \begin{bmatrix}
C\\
-C
    \end{bmatrix} \in \mathbb{R}^{2 n_G\times 2}
    \end{equation}
\begin{equation}
    b = A \cdot \centers + |{A G}| \; 1_{m\times 1} \in \mathbb{R}^{2 n_G}
\end{equation}
where $i = 1, \ldots, n_G$ indexes the number of generators. 
\end{prop}
In this work, zonotopes are used to describe the social reachable set for all agents, i.e., the ego-agent and pedestrians.
We seek to learn a sequence of social zonotopes for the ego-agent $\zonotope \ego \currq$, each of which contains two consecutive waypoints of the ego-agent's future social trajectory $\set{T} \ego \future$, thereby approximating the agent's continuous-time motion similar to \cite{paparussozapp}.
\begin{defn}[Social Zonotope $\zonotope \ego \currq$]
A social zonotope for the ego-agent's $q^{\rm th}$ step is $\zonotope \ego \currq =\zonofn{\centers \currq, G \currq}$, satisfying that the future traj ... $\set{T} \ego \future \in \bigcup\limits_{q=t}^{t_{f-1}} \zonotope \ego_{q}$.
\end{defn}

%% file: sections/06_learning_arch.tex
\section{Social Zonotope Network}
\label{sec:social_zono_net}
This section introduces the Social Zonotope Network (SZN) architecture and the loss functions used during training, which are designed both for shaping the social zonotopes and for ensuring the physical viability of the path for bipedal locomotion.
A salient feature of our social zonotope network is to learn the zonotope representation directly as an output of the neural network enabling real-time reachability-based planning and collision avoidance in the MPC introduced later in Sec.~\ref{sec:SDMPC}.

\subsection{Learning Architecture}
\label{subsec:learning_arch}

\begin{figure*}[t]
\centerline{\includegraphics[width=.95\textwidth]{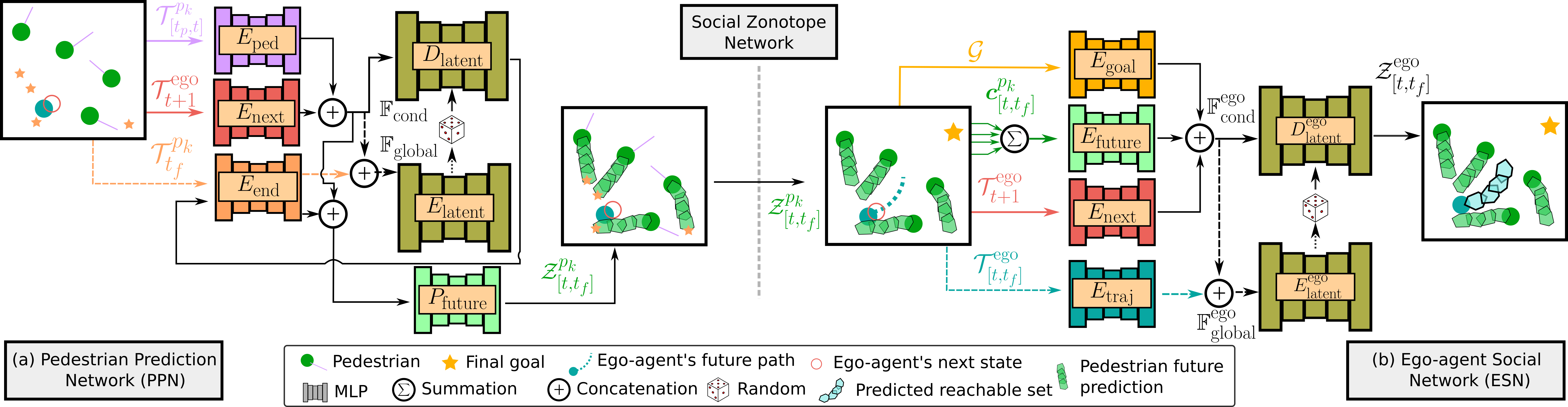}}
\caption{The Social Zonotope Network (SZN) is comprised of two coupled neural networks to both predict the reachable set of the surrounding pedestrians and learn the social reachable set of the ego-agent. (a) shows the pedestrian prediction network, conditioned on the pedestrian endpoints and the immediate change in the ego-agent's state. (b) shows the ego-agent social network conditioned on the pedestrians' future prediction, the immediate change in the ego-agent's state, and the ego-agent's goal location. Dashed connections are used during training only.}
\label{fig:arch}
\end{figure*}
We use a conditional variational autoencoder (CVAE) architecture to learn the ego-agent's future trajectory conditioned on the final destination goal, the immediate change in the ego-agent's state, and the surrounding pedestrians' past trajectories.
The proposed architecture uses Multi-Layer Perceptrons (MLP) with ReLU non-linearity for all the sub-networks.

Our Social Zonotope Network (SZN) is comprised of two coupled neural networks to not only predict the reachable set of the surrounding pedestrians but also learn the social reachable set of the ego-agent as shown in Fig.~\ref{fig:arch}.
\subsubsection{Pedestrian Prediction Network (PPN)} The pedestrian prediction network (shown in Fig.~\ref{fig:arch}(a)) is inspired by PECNet~\cite{mangalam2020not}, where the endpoint of the predicted pedestrian trajectory $\set{T} \ped_{t_f}$ ($t_f = 8$ indicating a 8-step horizon) is learned first, and then the future trajectory for the $[t, t_f]$ horizon is predicted.
Our proposed network deviates from PECNet in three ways.
First, the pedestrian future trajectory is also conditioned on the immediate change in the ego-agent's state $\set{T} \ego_{t+1}$ (shown in red in Fig.~\ref{fig:arch}(a)).
This coupling of the pedestrian prediction and ego-agent planning networks is intended to capture the effect of the ego-agent's control on the future trajectories of the surrounding pedestrians~\cite{schaefer2021leveraging, mavrogiannis2023core}. 
Second, the output of the network is the pedestrian's future reachable set parameterized as zonotopes $\zonotope \ped \future$  rather than point-based trajectories for robust collision checking and uncertainty parameterization~\cite{paparussozapp, selim2022safe, kousik2019safe}.
Third, we replace PECNet's social pooling module with a simple ego-agent sensory radius threshold for computational efficiency, and avoid the complexity when integrating our trained SZN into the MPC for a unified prediction and planning framework in Sec.~\ref{sec:SDMPC}.

The pedestrians' past trajectories $\set{T} \ped\past$ are encoded by a neural network $E_{\rm ped}$ as seen by the purple arrow in Fig.~\ref{fig:arch}(a), while the incremental change in the ego-agent state representing the ego-agent control is encoded by $E_{\rm next}$ as seen by the red arrow in Fig.~\ref{fig:arch}(a).
This allows us to condition the prediction of the pedestrians' trajectory on the ego-agent's control.
The resultant latent features $E_{\rm ped}(\set{T} \ped\past)$ and $E_{\rm next}(\set{T} \ego_{t +1})$ are then concatenated and used as the condition features $\mathbb{F}_{\rm cond}$.
The pedestrian's endpoint locations are encoded as $E_{\rm end}$ as seen by the orange arrows in Fig.~\ref{fig:arch}(a).
The resultant latent features $E_{\rm end}(\set{T} \ped_{t_f})$  are then concatenated with $\mathbb{F}_{\rm cond}$ as global features $\mathbb{F}_{\rm global}$ and encoded by the latent encoder $E_{\rm latent}$.
We randomly sample features from a normal distribution $\mathcal{N} (\boldsymbol{\mu},\boldsymbol{\sigma})$ generated by the $E_{\rm latent}$ module, and concatenate them with $\mathbb{F}_{\rm cond}$.
This concatenated information is then passed into the latent decoder $D_{\rm latent}$.
Then $D_{\rm latent}$ outputs the predicted endpoint that is passed again through $E_{\rm end}$.
The output is concatenated again with $\mathbb{F}_{\rm cond}$ and passed to another encoder $P_{\rm future}$ to output the predicted zonotopes of the pedestrians $\zonotope \ped \future$.

\subsubsection{Ego-agent Social Network (ESN)}

Our ESN architecture is shown in Fig.~\ref{fig:arch}(b).
The surrounding pedestrians' future zonotope centers $\centers\ped\future$ are aggregated through summation to take into account the collective effect of surrounding pedestrians\footnote{Other human trajectory learning modules include a social module to take into account the surrounding pedestrians effect such as social non-local pooling mask~\cite{mangalam2020not}, max-pooling~\cite{gupta2018social}, and sorting~\cite{sadeghian2019sophie}.}  while keeping a fixed-input-size architecture\cite{salzmann2020trajectron++, ivanovic2018generative, jain2016structural}.
The summed pedestrian features are then encoded by $E_{\rm future}$ as seen by the green arrows in Fig.~\ref{fig:arch}(b). 
The goal location for the ego-agent is encoded by $E_{\rm goal}$, while the incremental change in the ego-agent state is encoded by $E_{\rm next}$ as seen by the orange and red arrows respectively in Fig.~\ref{fig:arch}(b).
The resultant latent features $E_{\rm future}(\sum_{k=1}^{m}\centers\ped_{[t,t_f]})$, $E_{\rm goal}(\mathcal{G})$ and $E_{\rm next}(\set{T} \ego_{t +1})$ are then concatenated and used as the condition features $\mathbb{F}\ego_{\rm cond}$ for the CVAE.
The ground truth of the ego-agent's future trajectory $\set{T} \ego\future$ is encoded by $E_{\rm traj}$ as shown by the cyan arrows in Fig.~\ref{fig:arch}(b).
The resultant latent features $E_{\rm traj}(\set{T} \ego\future)$ are then concatenated with $\mathbb{F}\ego_{\rm cond}$ as global features $\mathbb{F}\ego_{\rm global}$ and encoded by the latent encoder $E\ego_{\rm latent}$.
Similarly, we randomly sample features from a normal distribution $\mathcal{N} (\boldsymbol{\mu},\boldsymbol{\sigma})$ generated by the $E\ego_{\rm latent}$ module, and concatenate them with $\mathbb{F}\ego_{\rm cond}$.
This concatenated information is then passed into the latent decoder $D\ego_{\rm latent}$, resulting in our prediction of the ego-agent's future reachable set $\zonotope \ego\future$.

\begin{rem}
    Including the $E_{\rm next}$ encoder in both the PPN and ESN facilitates seamless integration with a step-by-step MPC, as one of the MPC's decision variables, i.e., $\Delta\position\ego$, is used as inputs to $E_{\rm next}$ as detailed in Sec.~\ref{sec:SDMPC}.
\end{rem}


\subsection{Zonotope Shaping Loss Functions}
\label{subsec:zonotope_loss}
\begin{figure}[t]
\centerline{\includegraphics[width=.75\columnwidth]{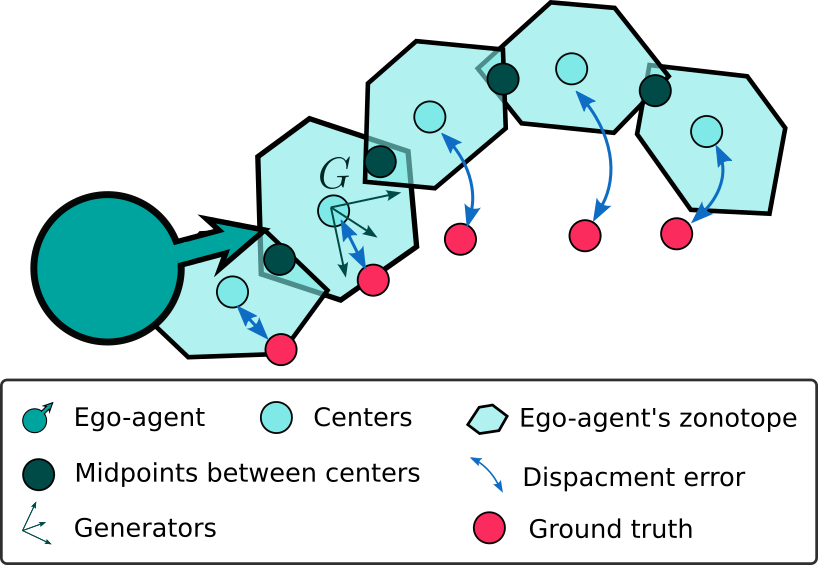}}

\caption{Our zonotope shaping loss functions. The loss aims to learn interconnected zonotopes that engulf the ground truth path.}
\label{fig:zonotope_shaping}
\end{figure}

We propose novel zonotope shaping loss functions for both PPN and ESN, since both networks output zonotopes.
These losses seek to achieve three goals: (i) penalize deviation of the centers of the zonotopes from the ground truth future trajectory; (ii) generate intersecting zonotopes between consecutive walking steps; and (iii) reduce the size of the zonotopes to avoid unnecessary, excessively large zonotopes.
Based on these goals, we implement the following losses to shape the output zonotopes (see Fig.~\ref{fig:zonotope_shaping}):
\begin{enumerate}
    \item Average displacement error between the predicted centers and midpoint of the ground truth trajectory  $\mathcal{T}_{{\rm mid},i}$:
    \begin{equation*}
        \loss_{\rm ADE} = \frac{\sum_{i=1}^{t_{f}-1}\|\mathcal{T}_{{\rm mid},i} - \centers_i \|}{t_{f}-1}
    \end{equation*}
    \item Final displacement error between the last predicted center and the final midpoint of the ground truth trajectory:
    \begin{equation*}
        \loss_{\rm FDE} = \| \mathcal{T}_{{\rm mid}, t_{f}-1} - \centers_{t_{f}-1} \|
    \end{equation*}
    
    \item The midpoint between the current center and previous center $\centers^p_{{\rm mid},i}$ is contained in the current zonotope:
    \begin{equation*}
        \loss_{\rm prev} = \sum_{i=0}^{t_{f}-1} {\rm ReLU}(A_i \cdot \centers_{{\rm mid},i-1} - b_i)
    \end{equation*}
    where $\centers_{{\rm mid},-1}$ is the initial point of the ground truth trajectory $\set{T}\future$, i.e., the current location of the ego-agent.
    \item The midpoint between the current center and the next center $\centers^n_{{\rm mid},i}$ is contained in the current zonotope:
    \begin{equation*}
        \loss_{\rm next} =  \sum_{i=0}^{t_{f}-1} {\rm ReLU}(A_i \cdot \centers_{{\rm mid},i+1} - b_i)
    \end{equation*}
    where $\centers_{{\rm mid},t_f}$ is the endpoint of the ground truth trajectory $\set{T}\future$.
    \item Regulating the size of the zonotope, by penalizing the norm of the generators such that the neural network does not produce excessively large zonotopes that contain the ground truth trajectory:
    \begin{equation*}
        \loss_{G} =   \sum^{n_G}_{i=1}\| l_{G}[i]- d_i \|
    \end{equation*} 
\end{enumerate}
where $d_i$ is the desired length for each $i^{\rm th}$ generator.
We sum the zonotope shaping losses listed above into the total loss term $\loss_{\zonotope }$.
Similar to PECNet ~\cite{mangalam2020not}, we use Kullback–Leibler divergence to train the output of the latent encoder, aiming to regulate the divergence between the encoded distribution $\mathcal{N} (\boldsymbol{\mu},\boldsymbol{\sigma})$and the standard normal distribution $\mathcal{N} (0,\boldsymbol{I})$:
\begin{equation*}
    \loss_{\rm KL} = D_{\rm KL}\big(\mathcal{N} (\boldsymbol{\mu},\boldsymbol{\sigma})\ \|\ \mathcal{N} (0,\boldsymbol{I})\big)
\end{equation*}
Next, we introduce robot-specific losses for ESN to promote locomotion safety and ensure that the Digit robot is able to reach consecutive zonotopes in consecutive walking steps. 
\subsection{
Incorporating Robot Safety Specifications}
\label{subsec:stl_loss}

To deploy the learning-based social path planner on the Digit robot, it is essential to consider the features of bipedal locomotion such as kinematic and dynamic constraints.
To this end, we introduce additional losses to ensure the learned path is viable for bipedal locomotion.

Signal Temporal Logic (STL) is a well-established temporal logic language to formally encode natural language into mathematical representation for control synthesis \cite{maler2004monitoring}.
More importantly, the quantitative semantics of STL offer a measure of the robustness of an STL specification $\rho (s_t,\phi)$, i.e., quantify the satisfaction or violation of the specification $\phi$ given a specific signal $s_t$.
Positive robustness values, i.e., $\rho (s_t,\phi) > 0$, indicate specification satisfaction, while negative robustness values indicate a violation.
The authors in \cite{leung2021back} present STLCG, a tool that transforms STL formulas into computational graphs to be used in gradient-based problems such as neural network learning.
To this end, we leverage a similar technique to formally incorporate desired locomotion safety behaviors into our learning framework by encoding STL specifications as additional loss functions that penalize STL formula violation \cite{leung2021back, li2021vehicle}.

The locomotion specifications are derived based on our previously introduced Reduced-Order Model (ROM) safety theorems \cite{shamsah2023integrated} and our empirical knowledge about the locomotion safety of Digit \cite{agility} during our experiments.
To maintain balance, the ROM Center of Mass (CoM) velocity ought to be bounded based on the step length and heading change~\cite{shamsah2023integrated}.
Therefore, we design STL specifications to regulate $\centers\ego\future$ to limit the sagittal and lateral COM velocities as well as the heading change between consecutive walking steps, based on prespecified thresholds. 

\subsubsection{Locomotion Velocity Specification $\phi_{\rm vel}$}
Let $s^{v_{\rm sag}}_{[t+1, t_f]}$ and $s^{v_{\rm lat}}_{[t+1, t_f]}$ be signals representing the velocity of $\centers\ego\future$ (via finite difference) in the sagittal and lateral directions, respectively.
The locomotion velocity specification has:
\begin{align}\begin{split}
\phi_{\rm sag} &= \square_{[t+1, t_f]}(s^{v_{\rm sag}}_{[t+1, t_f]} \leq v_{\rm max} \wedge s^{v_{\rm sag}}_{[t+1, t_f]} \geq v_{\rm min} ), \\
\phi_{\rm lat} &= \square_{[t+1, t_f]}(s^{v_{\rm lat}}_{[t+1, t_f]} \leq v_{\rm lat} \wedge s^{v_{\rm lat}}_{[t+1, t_f]} \geq - v_{\rm lat} ), \\
\phi_{\rm vel} &=  \phi_{\rm sag} \wedge  \phi_{\rm lat},
\end{split}\end{align}
where $\square_{[a,b]}(c)$ denotes that specification $c$ must be satisfied for all times $t \in [a,b]$.

Accordingly, the loss of the locomotion velocity specification is defined as:
\begin{equation}
    \loss_{\phi_{\rm vel}} = \underbrace{{\rm ReLU}(-\rho((s^{v_{\rm sag}},s^{v_{\rm lat}}),\phi_{\rm vel}))}_{\textit{velocity violation}}
    \label{eq:velocity}
\end{equation}

\subsubsection{Heading Change Specification $\phi_{\Delta \theta}$}
Let $s^{\Delta \theta}_{[t+1]}$ be a signal equal to the heading change between $\centers\ego_{t}$ and $\centers\ego_{t+1}$. The heading change specification is:
\begin{equation*}
\phi_{\Delta \theta} = \square_{[t+1, t_f]}(s^{\Delta \theta}_{[t+1, t_f]} < \Delta \theta_{\rm max} \wedge s^{\Delta \theta}_{[t+1, t_f]} > -\Delta \theta_{\rm max}) 
\end{equation*}

Therefore, the loss of the heading change specification is:
\begin{equation}
    \loss_{\phi_{\Delta \theta}} = \underbrace{{\rm ReLU}(-\rho(s^{\Delta \theta},\phi_{\Delta \theta}))}_{\textit{heading change violation}}
    \label{eq:heading_change}
\end{equation}

We sum the STL locomotion losses into a single term $\loss_{\rm STL}$.
The network is trained end to end using the following loss function: 
\begin{align}\loss =\alpha_1 \loss_{KL} + \alpha_2 \loss_{\zonotope } + \alpha_3 \loss_{\rm STL}
\end{align}
where these $\alpha$ parameters are  weighting coefficients. 

%% file: sections/07_zonotope_refinement.tex
\section{Zonotope Refinement for Social Acceptability  and Uncertainty Parameterization }\label{sec:zonotope_refinement}
Zonotopes have a desirable property of allowing computationally efficient downstream online refinements. 
In this section, we introduce two types of zonotope refinements for the ego-agent based on (1) personal space preference for increased social acceptability, and (2) modeling error compensation of robot dynamics discrepancy between the ROM and the full-order model of our Digit robot.

\begin{rem}
    Our key insight is that, by using zonotopes as the output format for our neural networks, we can easily postprocess the network outputs to incorporate sources of error that would be either difficult for the neural network to learn directly, or may change at runtime when the network cannot be retrained.
\end{rem}

\subsection{Personal Space Modulation}\label{subsec:personal_space}
Another consideration for social acceptability is the personal space surrounding every agent~\cite {hall1969hidden}. We leverage the adaptable nature of zonotope parameterization, which can be modulated in specific directions, to develop personal space generators.
These generators are inspired by the sociological study of proxemics~\cite{hall1969hidden} and its application to social navigation~\cite{patompak2020learning, rios2015proxemics}. 
The personal space generator matrix, ${G}^{\rm PS}$, is formulated as follows:
\begin{equation}
 {G}^{\rm PS}=\text{diag}(a, b) \cdot \begin{bmatrix} \cos(\theta) & -\sin(\theta) \\ \sin(\theta) & \cos(\theta) \end{bmatrix}
\end{equation}
The parameters $a$ and $b$ represent the scalar distances that define the personal space along the sagittal and lateral axes, respectively.
These distances are rotated by the angle $\theta$ to align the personal space with the ego-agent's current walking orientation in both directions as shown in Fig.~\ref{fig:zono_refine}(a).
The new ego-agent's zonotope is then augmented as $\hat{\set{Z}} \ego =\mathscr{Z}(\boldsymbol{c} \ego, [{G}^{{\rm ego}} \;{G}^{\rm PS}])$.

\subsection{Robot Modeling Error Compensation}
\label{subsec:Modelling_error_refinements}

Using zonotopes as a representation of reachable sets allows for online compensation for the robot dynamics discrepancy between the ROM and the full-order model of our Digit robot.

\begin{figure}[t]
\centerline{\includegraphics[width=.8\columnwidth]{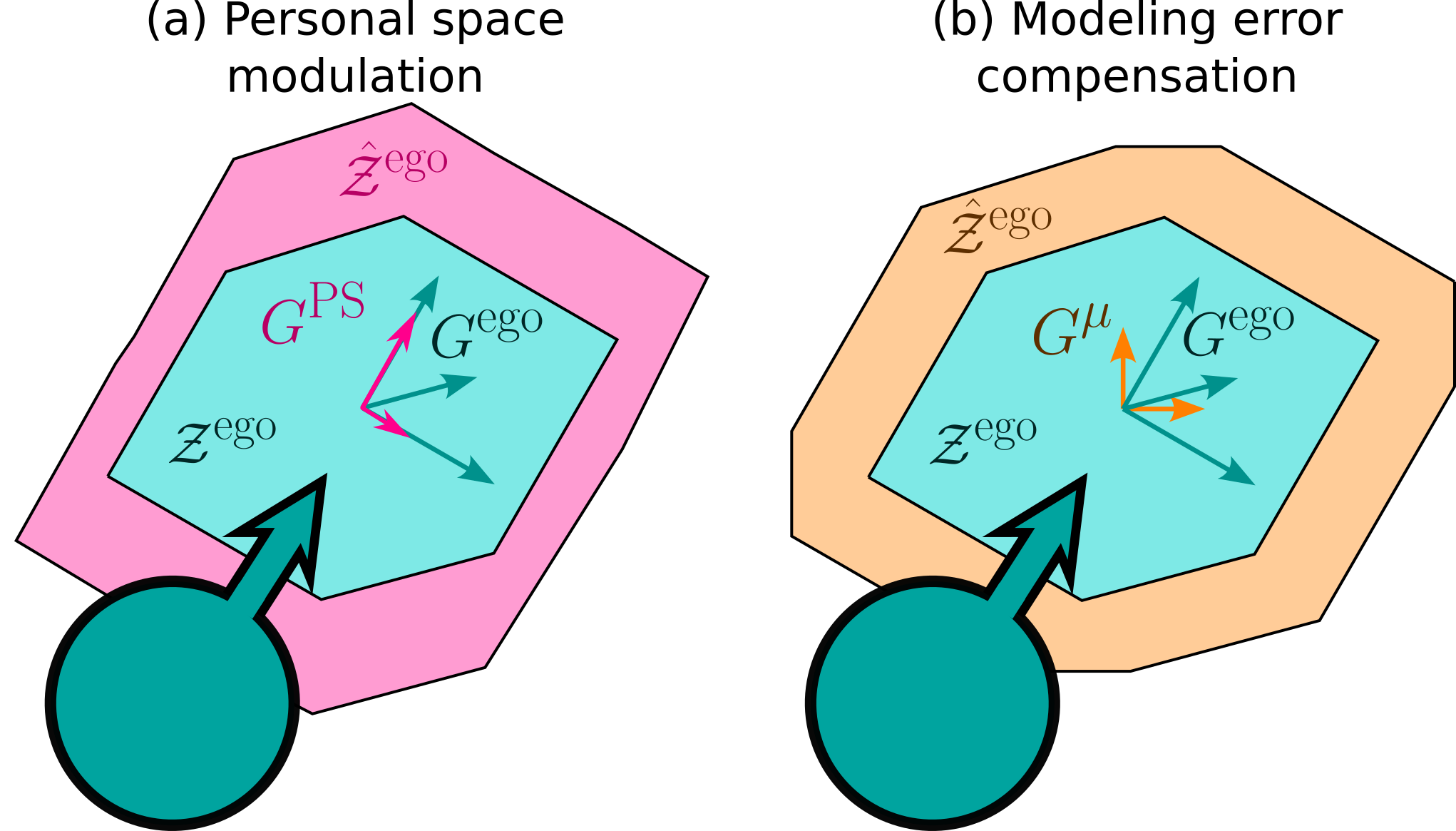}}
\caption{Two types of zonotope refinements: (a) zonotope refinement based on personal space preference for improved social acceptability, where the generators in $G^{\rm PS}$ are in the local sagittal and lateral directions as shown by the pink arrows. (b) zonotope refinement based on a learned GP model of the model discrepancy between ROM and full-order models, where the generators in $G^{\mu}$ are in the global $x$ and $y$ directions as shown by the orange arrows.}
\label{fig:zono_refine}
\end{figure}

We learn the modeling errors using Gaussian Process (GP) regression\cite{jiang2023abstraction, muenprasitivej2024bipedal}.
We introduce a GP\footnote{The GP model is trained offline.} that takes in the current sagittal velocity of the robot $v\local\currq$ and the MPC previous optimal solution for $(v\local_*, u^{\Delta \theta}_{*})$. To characterize the discrepancy, we chose these parameters as representative state variables because they effectively represent the key parameters contributing to the discrepancy between the ROM and low-level ALIP controller~\cite{Gong2022AngularMomentum}. The GP model then outputs the expected mean deviation $\boldsymbol{\mu}=(\mu_x, \mu_y)$ and variance $\boldsymbol{\sigma}=(\sigma^2_x, \sigma^2_y)$ in robot's Euclidean position at the next walking step (See Fig.~\ref{fig:block_framework}).
The Gaussian mean from the GP model is used to design a new generator matrix $G^{\mu} = \text{diag}(\mu_x, \mu_y)$ such that the ego-agent's zonotope becomes $\hat{\set{Z}} \ego \nextq =\mathscr{Z}(\boldsymbol{c} \ego, [{G}^{{\rm ego}} \; {G}^{\rm PS} \; G^{\mu}])$ to compensate for the anticipated mismatch between the ROM dynamics and the full-order dynamics (see Fig.~\ref{fig:zono_refine}(b).

%% file: sections/08_MPC.tex
\section{Social MPC} \label{sec:SDMPC}

To safely navigate the human-crowded environment we propose to solve the following trajectory optimization problem that encodes the SZN in the previous section as constraints:
\begin{subequations}
\label{general_mpc}
\begin{align}
\min_{X, U} \quad &\sum_{q = 0}^{N-1} J(\state_q, \ctrl_q) + J_N(\state_{N}) \label{general_mpc_cost} \\
\textrm{s.t.} \quad & \state \nextq = \Phi(\state \currq, \ctrl \currq), \forall q \label{general_mpc_dynamics}\\
    & \state_0 = \state_{\rm init}, \;  (\state \currq, \ctrl \currq) \in \set{XU} \currq, \forall q    \\
  & \position\ego \nextq \in \hat{\set{Z}} \ego \nextq(\Delta\position\ego_{q}, \set{E} \currq) , \forall q\label{general_mpc_stay_within}\\
  & \hat{\set{Z}} \ego \nextq(\Delta \position\ego_{q} , \set{E} \currq)  \bigcap \set{Z}^{{p_{k \currq}}} \nextq = \emptyset, \; \forall \; q, k \currq \label{general_mpc_avoid}
\end{align}
\end{subequations}
where the decision variables include a state sequence $X = \{\state_1, \ldots, \state_{N}\}$ and a control sequence $U = \{\ctrl_1, \ldots, \ctrl_{N-1} \}$, the running and terminal costs~\eqref{general_mpc_cost} are designed to reach the goal and promote social acceptability, subject to the ROM dynamics~\eqref{general_mpc_dynamics} (Sec.~\ref{subsec:ROM}).
Constraint~\eqref{general_mpc_stay_within} requires the ego-agent at the next $(q+1)^{\rm th}$ walking step to stay within the reachable set, while constraint~\eqref{general_mpc_avoid} requires the ego-agent to avoid collision with the pedestrians. $\set{E} \currq = (\set{T}\ped\pastq,\set{G})$ denotes the environment state at the $q^{\rm th}$ walking step.
Next, we will introduce the kinematics constraints in Sec.~\ref{subsec:kin_const}, and navigation constraints in Sec.~\ref{subsec:reachability_and_nav_const}, social acceptability cost function (Sec.~\ref{subsec:cost}), and finally reformulate the MPC in~(\ref{general_mpc}) with a version for implementation (Sec.~\ref{subsec:MPC}).

\subsection{Kinematics Constraints}
\label{subsec:kin_const}
To prevent the LIP dynamics from taking a step that is kinematically infeasible by the Digit robot, we implement the following constraint
\begin{equation}
    \set{XU} \currq = \{(\state \currq, \ctrl \currq) \; | \; \state_{\rm lb} \leq \state \currq \leq \state_{\rm ub} \; \text{and} \; \ctrl_{\rm lb} \leq  \ctrl \currq \leq \ctrl_{\rm ub} \}
    \label{eq:const}
\end{equation}

where $\state_{\rm lb}$ and $\state_{\rm ub}$ are the lower and upper bounds of $\state \currq$ respectively, and $\ctrl_{\rm lb}$ and $\ctrl_{\rm ub}$ are the bounds for $\ctrl \currq$ (See Fig.~\ref{fig:LIP}).

\subsection{Reachability and Navigation Safety Constraints}
\label{subsec:reachability_and_nav_const}
To enforce navigation safety (i.e., collision avoidance), we require that Digit remains in the social zonotope $\set{Z} \ego$ and outside of the surrounding pedestrians reachable set $ \set{Z}^{{p_{k}}}$.
\subsubsection{Reachability constraint}

For the robot's CoM to remain inside the desired zonotope for the next walking step $\hat{\set{Z}} \ego \nextq$, we represent the zonotope using half-space representation as shown in Prop.~\ref{prop:half_space}.
The constraint is reformulated as such:
\begin{equation}
  \max(\hat{A} \ego \position \ego - \hat{b} \ego) \leq 0
\end{equation}

\subsubsection{Navigation Safety Constraint}
For pedestrian collision avoidance, we require that the reachable set of the ego-agent does not intersect with that of the pedestrians for each walking step.
Therefore, we design a new zonotope for the ego-agent as Minkowski sum of the ego-agent's zonotope and the pedestrian's zonotope centered around the ego-agent $\set{Z} \mink =\mathscr{Z}(\boldsymbol{c} \ego, [{G}^{{\rm ego}} \; {G}^{\rm PS} \; G^{\mu} \; G \ped])$ to check for collision with the pedestrians' zonotope following Prop.~\ref{prop:intersection}.
We then represent $\set{Z} \mink$ using a half-space representation parametersized by $A_k\mink$ and $b_k\mink$, as per Prop.~\ref{prop:half_space}, and require that the pedestrian position is outside the Minkowski-summed zonotope.
Thus, for each $k$\textsuperscript{th} pedestrian, we have the following constraint:
\begin{equation}
    \max(A_k \mink \position_k - b_k \mink) > 0.
\end{equation}

\subsection{Cost Function and Social Acceptability Metric}
\label{subsec:cost}

\begin{figure}[t]
\centerline{\includegraphics[width=.95\columnwidth]{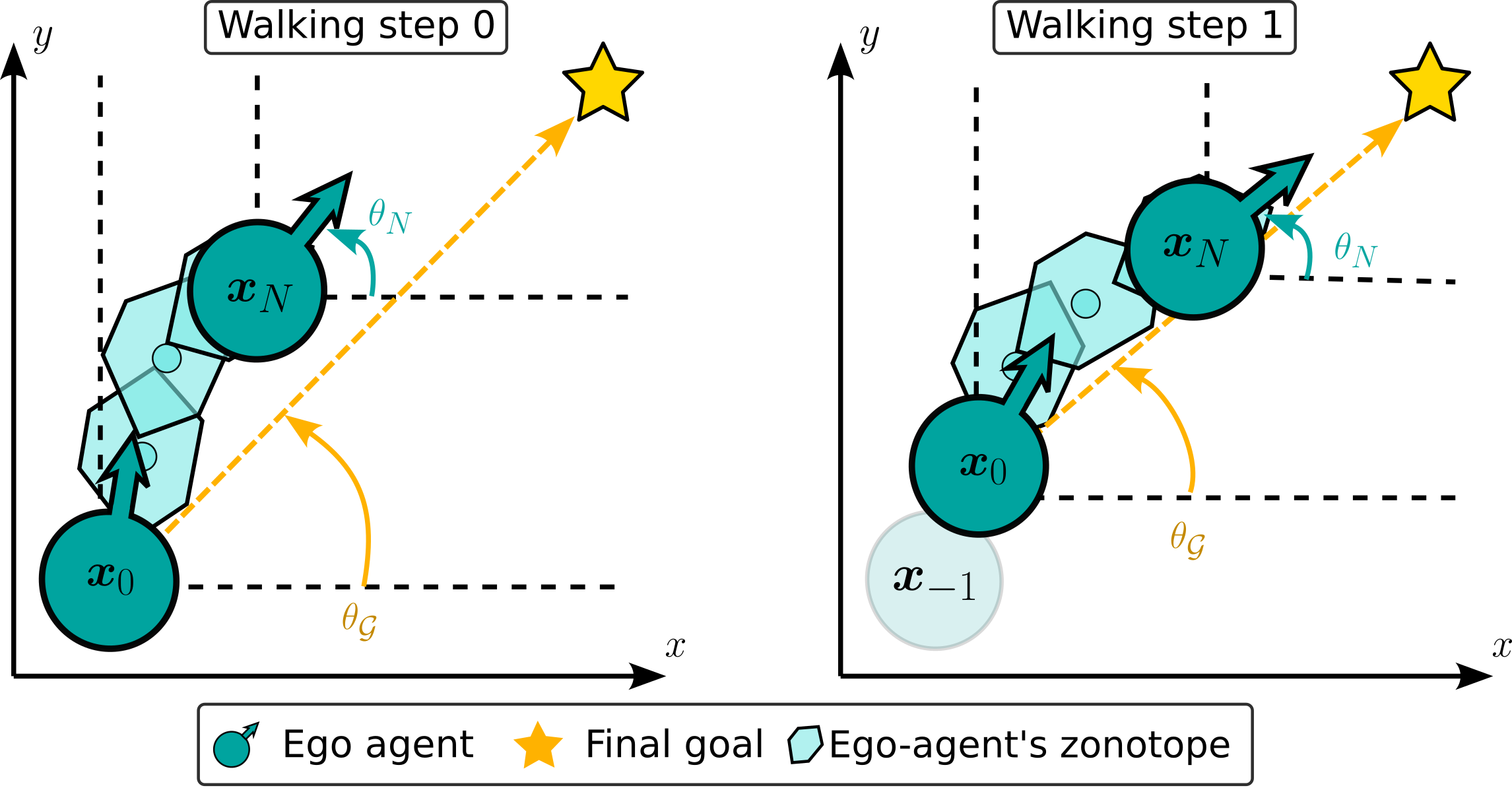}}
\caption{Illustration of the terminal goal angle $\theta_\set{G}$, a state dependent on the initial state $\state_0$. $\theta_\set{G}$ is designed to guide the ego-agent towards the final goal location. SZN-MPC optimizes over $N$ walking step horizon such that the ego-agent heading at the end of the planning horizon $\theta_N$ aligns with $\theta_\set{G}(\state_0)$. After executing a walking step (see the figure on the right), we update $\theta_\set{G}(\state_0)$ based on the new initial state $\state_0$ for the ego-agent.}
\label{fig:terminal_cost}
\end{figure}

\begin{figure}[t]
\centerline{\includegraphics[width=.95\columnwidth]{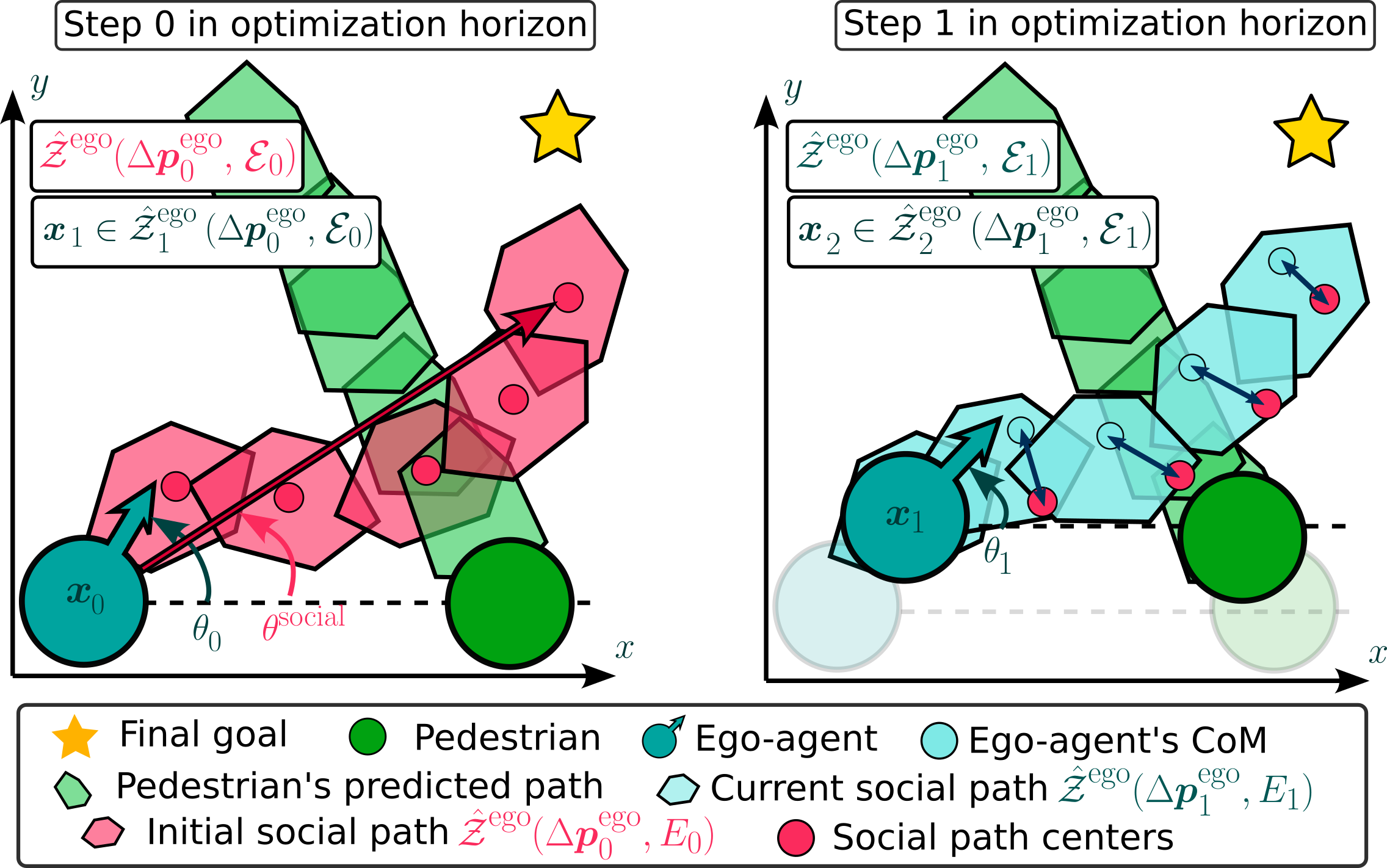}}
\caption{Illustration of the social acceptability cost. In the initial step in the optimization horizon (figure on the left), the learned social path $\hat{\set{Z}} \ego(\Delta\position\ego_{0}, \set{E}_0)$ is shown in the red zonotopes, and the active constraint is $\state_1 \in \hat{\set{Z}} \ego_1(\Delta\position\ego_{0}, \set{E}_0)$. In the following step in the optimization horizon (figure on the right) the learned social path $\hat{\set{Z}} \ego(\Delta\position\ego_{1}, \set{E}_1)$ (shown by the cyan zonotopes) is based on the \textit{current} environment state $\set{E}_1$ and ego-agent \textit{current} location $\position\ego_{1}$, and the active constraint is $\state_{2} \in \hat{\set{Z}} \ego _{2}(\Delta\position\ego_{1}, \set{E}_1)$. The social acceptability cost aims to minimize the difference between the centers $\hat{\boldsymbol{c}}^{\rm social}_q$ (red dots) of $\hat{\set{Z}}_{q} \ego(\Delta\position\ego_{0}, \set{E}_0) \; \forall q$ and the ego-agent ROM CoM $\position \ego_q$ (cyan dots). The arrows between the social path centers $\hat{\boldsymbol{c}}^{\rm social}$ and ego-agent ROM CoM $\position \ego$ indicate the distance that the social acceptability cost aims to minimize $\| \hat{\boldsymbol{c}}^{\rm social} \currq -\position \ego_q \|$. }
\label{fig:social_path}
\end{figure}
The MPC cost function is designed to drive the CoM state to a goal location $\set{G}$ and to promote social acceptability.
The terminal cost penalizes (1) the distance between the current ROM state and the global goal state $\set{G}$ in the 2D world coordinate, and (2) the ego-agent heading angle deviation from the heading angle pointing toward the final goal location (see Fig.\ref{fig:terminal_cost}) to avoid abnormal walking gaits, such as the robot moving toward the goal but in a way of heading opposite to the goal location while walking backward.
\begin{equation}
    J_N(\state_{N}) = \| \state_{N} - \state_{\set{G}} \|^2_{W_1} + \| \theta_{N} - \theta_{\set{G}}(\state_0) \|^2_{W_2}
\end{equation}
where $\state_{\set{G}} = (\set{G}, v_{\rm terminal})$, $\set{G}=(x^{\rm dest}, y^{\rm dest})$, and $\theta_\set{G}(\state_0)$ is the angle between the ego-agent's current position and the final goal location $\set{G}$ (see more details in Fig.~\ref{fig:terminal_cost}).

The MPC constrains the ROM CoM $\position\ego_q$ to stay within the ego-aganet's zonotope~\eqref{general_mpc_stay_within} and avoid collisions~\eqref{general_mpc_avoid}.
These constraints, along with the changes in the environment $(\set{E}\currq\neq \set{E}\nextq, \forall q)$, might cause the generated ROM CoM $\position\ego_q$ trajectory to deviate from the learned social path $\hat{\set{Z}}_{q+1} \ego(\Delta\position\ego_{0}, \set{E}_0)$ with the centers $\hat{\boldsymbol{c}}^{\rm social}_q, \forall q$. This social path is generated from the \textit{initial} environment $ \set{E}_0$.

Therefore, we incorporate a social acceptability metric by creating a cost that penalizes deviation of the ROM CoM $\position\ego_q$ from the learned social path $\hat{\set{Z}} \ego\nextq(\Delta\position\ego_{0}, \set{E}_0)$ as shown in Fig.~\ref{fig:social_path}.
We set the social acceptability metric as: (1) the distance between the ROM CoM $\position\ego_q$ and the centers of the learned social path $\hat{\boldsymbol{c}}^{\rm social}_q$, and (2) the difference between the ego-agent current heading $\theta_q$ and social heading angle $\theta^{\rm social}$, i.e., the angle between the ego-agent's initial position and the $\hat{\boldsymbol{c}}^{\rm social}_N$, as shown in 
Fig.~\ref{fig:social_path}.
Thus, we set the running cost of social acceptability as follows:
\begin{equation}
    J_{\rm social}(\state \currq) = \| \hat{\boldsymbol{c}}^{\rm social} \currq -\position \ego_q \|^2_{W_3} + \| \theta^{\rm social} - \theta_{q}\|^2_{W_4}
    \label{eq:socia_cost}
\end{equation}

Including such social acceptability metric as a cost function, will guide SZN-MPC to generate the CoM trajectory that (1) tracks the learned social path $\hat{\set{Z}} \ego\nextq(\Delta\position\ego_{0}, \set{E}_0)$, (2) is within the next zonotope based on the current environment state $\set{E}_q$, and (3) is collision-free (i.e., constraints ~\eqref{general_mpc_stay_within}-\eqref{general_mpc_avoid}).

\begin{rem}
    The initial output of the neural network $\hat{\set{Z}} \ego\nextq(\Delta\position\ego_{0}, \set{E}_0)$ is not guaranteed to be collision-free at every walking step in the planning horizon. Therefore, the assumed socially acceptable path is set as a cost and not a constraint, to prioritize safety over social acceptability. 
\end{rem}

\subsection{Social Zonotope Network MPC Formulation}
\label{subsec:MPC}

We reformulate our Social Zonotope Network MPC (SZN-MPC) shown in~(\ref{general_mpc}) based on the aforementioned costs and constraints for implementation as follows:
\begin{subequations}
\label{eq:problem}
\begin{align}
\min_{X, U} \quad \sum_{q=0}^{N-1} &  J_{\rm social}(\state \currq)+  J_N(\state_{N})  \\
\textrm{s.t.} \quad & \state \nextq = \Phi(\state \currq, \ctrl \currq), \forall q\\
    & \state_0 = \state_{\rm init}, \;  (\state \currq, \ctrl \currq) \in \set{XU} \currq, \forall q   \label{eq:imp_kin_contd} \\
  &  \max(\hat{A} \ego \nextq \position \ego \nextq - \hat{b} \ego \nextq) \leq 0, \forall q  \label{eq:HS_mpc_reach}\\
  &  \max(A \mink \nextq \position_{k \nextq} - b \mink \nextq) > 0, \; \forall \;q, k \currq  \label{eq:HS_mpc_avoid}
\end{align}
\end{subequations}
where the kinematics constraint in~\eqref{eq:imp_kin_contd} is implemented using the local dynamics $(\Delta x \local, v\local)$. We provide detailed implementation parameters in Table~\ref{tab:params}.

%% file: sections/09_implementation.tex
\section{Results}\label{sec:results}
In this section, we present comprehensive results and their analysis of three distinct experiments designed to evaluate various aspects of our framework. The first experiment assesses the social acceptability and locomotion feasibility of the SZN-MPC. The second experiment benchmarks our SZN-MPC with a baseline approach for the LIP-based MPC that uses a discrete-time control barrier function (DCBF-MPC) for dynamic obstacle collision avoidance and trajectory planning. The final experiment tests the feasibility of implementing SZN-MPC on our Digit robot hardware. The section starts with implementation details.

\subsection{Implementation Details}
\label{sec:implementation}

\subsubsection{Training}
The social path planner module introduced in Sec.~\ref{sec:social_zono_net} was trained on the UCY~\cite{lerner2007crowds} and ETH~\cite{pellegrini2009walk} crowd datasets with the common leave-one-out approach, reminiscent of prior studies~\cite{salzmann2020trajectron++,mangalam2020not, gupta2018social, li2019conditional}. More specifically, we excluded the UNIV dataset from the training examples, and used it for testing. To evaluate the performance of incorporating robot-locomotion-specific STL specifications into the training, we trained two neural network models with and without the added robot-specific losses introduced in Sec.~\ref{subsec:stl_loss}. 
We employ a historical trajectory observation $\mathcal{T}^{p_k}_{[-8, 0]}$ for all neighboring pedestrians that are within a radius of $4$ m and a prediction horizon $\hat{\mathcal{T}}^{\rm ego}_{[0, 8]}$. For both pedestrians and the ego-agent, the duration of $8$ timesteps takes $3.2$ s ($8 \times 0.4$ s $= 3.2$ s). The network architecture details are shown in Appendix~\ref{app:network_arch}. The SZN is implemented and trained using PyTorch~\cite{paszke2019pytorch}.

\subsubsection{Pedestrian Simulation}
We utilize SGAN (Social Generative Adversarial Network)~\cite{gupta2018social}, a state-of-the-art human trajectory prediction model, for simulating pedestrians during our testing.
SGAN is specifically designed to grasp social interactions and dependencies among pedestrians.
It considers social context, including how people influence each other and move in groups.
This is important for creating realistic simulations of pedestrian motion.
Employing a prediction model different from our model ensures a fair evaluation by eliminating any inherent advantages of our proposed method\cite{schaefer2021leveraging}. In our simulation framework, SGAN incorporates the historical trajectories of both the pedestrians and the ego-agent, where the ego-agent is also treated as a pedestrian. This approach enhances the realism of the simulation by accounting for the interaction between the ego-agent and pedestrians in the environment.

\subsubsection{Testing Environment Setup}
\label{subsubsec:test_env}
The environment for all the following simulation tests is an open space of $14 \times 14$ m$^2$, with randomly generated pedestrians' initial trajectory.
The goal location is $\mathcal{G}=(6,12)$ m, and the ego-agent starting position is uniformly sampled along the $y$-axis as such $\boldsymbol{x}_0 = (0, \mathcal{U}_{[0, 13]},0)$ with $\theta_0 = 0$.
The MPC is solved with a planning horizon of $N=4$, SZN-MPC parameters are included in Table.~\ref{tab:params}.
Simulations and training are conducted using a 16-core Intel Xeon W-2245 CPU and an RTX-5000 GPU with 64 GB of memory. The SZN-MPC is implemented using do-mpc Python libraray~\cite{fiedler2023mpc} and CasADi~\cite{andersson2019casadi}. Digit is simulated using the MuJoCo simulator provided by Agility Robotics~\cite{agility} and visualized using  Nvidia Isaac Gym~\cite{makoviychuk2021isaac}, which allows an animation of pedestrian characters in the environment.

\begin{table}
    \centering
    \caption{SZN-MPC Parameters}
    \begin{tabular}{|c|c||c|c|}
    \hline
        parameter & value & parameter & value \Tstrut\Bstrut \\ [3pt]
    \hline
    \hline
    $ v\local_{\rm ub}$ & $1.0$ m/s &
    $ v\local_{\rm lb}$ & $-0.1$ m/s \Tstrut\Bstrut\\ [3pt]
    \hline
    $\Delta x \local_{\rm ub}$ & $0.2$ m &
    $\Delta x \local_{\rm lb}$ & $-0.2$ m \Tstrut\Bstrut \\ [3pt]
    \hline
       $u^{\Delta \theta}_{\rm ub}$ & $15^\circ$ & $u^{\Delta \theta}_{\rm lb}$ & $-15^\circ$ \Tstrut\Bstrut \\ [3pt]
    \hline
        $u^{f}_{\rm ub}$ & $0.4$ m & $u^{f}_{\rm lb}$ & $-0.1$ m \Tstrut\Bstrut \\ [3pt]
    \hline
       $d_1$  & $0.1$ & $d_2$ & $0.005$ \Tstrut\Bstrut \\ [3pt]
    \hline
        $v_{\rm terminal}$ & $0$ m/s & $n_{G}$ & $4$ \Tstrut\Bstrut \\ [3pt]
    \hline
        $W_1$ & $3$ & $W_2$ & $1$ \Tstrut\Bstrut \\[3pt]
    \hline
     $a$ & $0.3$ m & $b$ & $0.2$ m \Tstrut\Bstrut \\[3pt]
    \hline
     $\alpha_1$, $\alpha_3$ & $1$ & $\alpha_2$ & $100$ \Tstrut\Bstrut \\[3pt]
    \hline
    \end{tabular}
    
    \label{tab:params}
\end{table}

%% file: sections/10_results.tex
\subsection{Experiment 1: Social Acceptability and Locomotion Feasibility}
In this section, we aim to quantify the social acceptability of the generated paths of SZN, and the feasibility of the generated path for bipedal locomotion. We quantify the social acceptability using two methods: First, we measure the average displacement error (ADE) and final displacement error (FDE) of ESN's planned path compared to the ground truth; Second, we measure the social acceptability metric running SZN-MPC with and without the social acceptability cost function. For locomotion feasibility, we compare the tracking performance of the ROM to the social path with and without the locomotion losses introduced in Sec.~\ref{subsec:stl_loss}.

\subsubsection{Social Acceptability}
\begin{figure}[t]
\centerline{\includegraphics[width=.95\columnwidth]{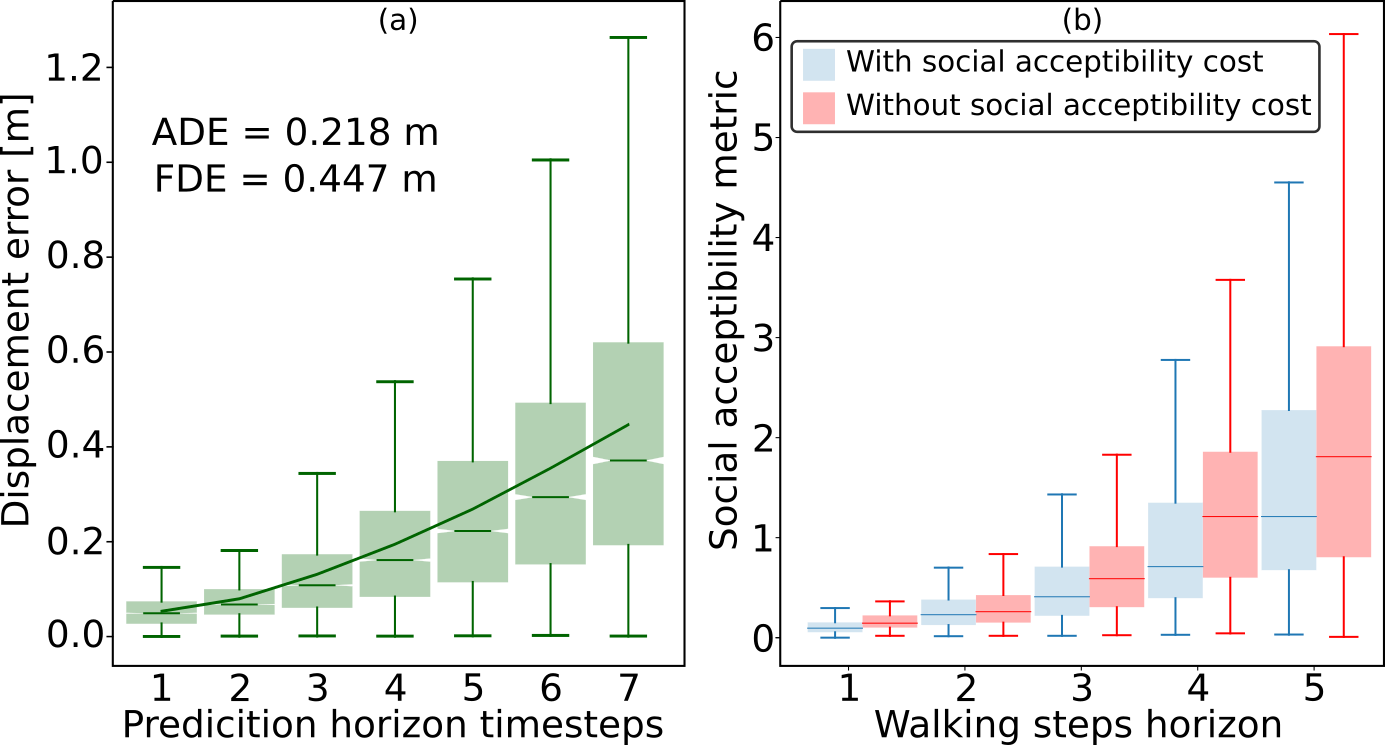}}
\caption{Quantitative results of social acceptability: (a) Show the displacement error between the prediction of ESN $\boldsymbol{c}^{\rm ego}$ and the ground truth data $\mathcal{T}^{\rm ego}_{\rm mid}$. The data is collected based on the UNIV dataset with $7831$ unique frames. The solid line shows the average displacement error at each prediction horizon. (b) shows the social acceptability metric for $5$ different trials with $30$ pedestrians and $100$ walking steps in each trial. The social acceptability metric is reduced when Problem~(\ref{eq:problem}) is solved with the social cost (\ref{eq:socia_cost}), thus yielding a socially acceptable path.}
\label{fig:quan_social_metric}
\end{figure}

\begin{figure}[t]
\centerline{\includegraphics[width=.95\columnwidth]{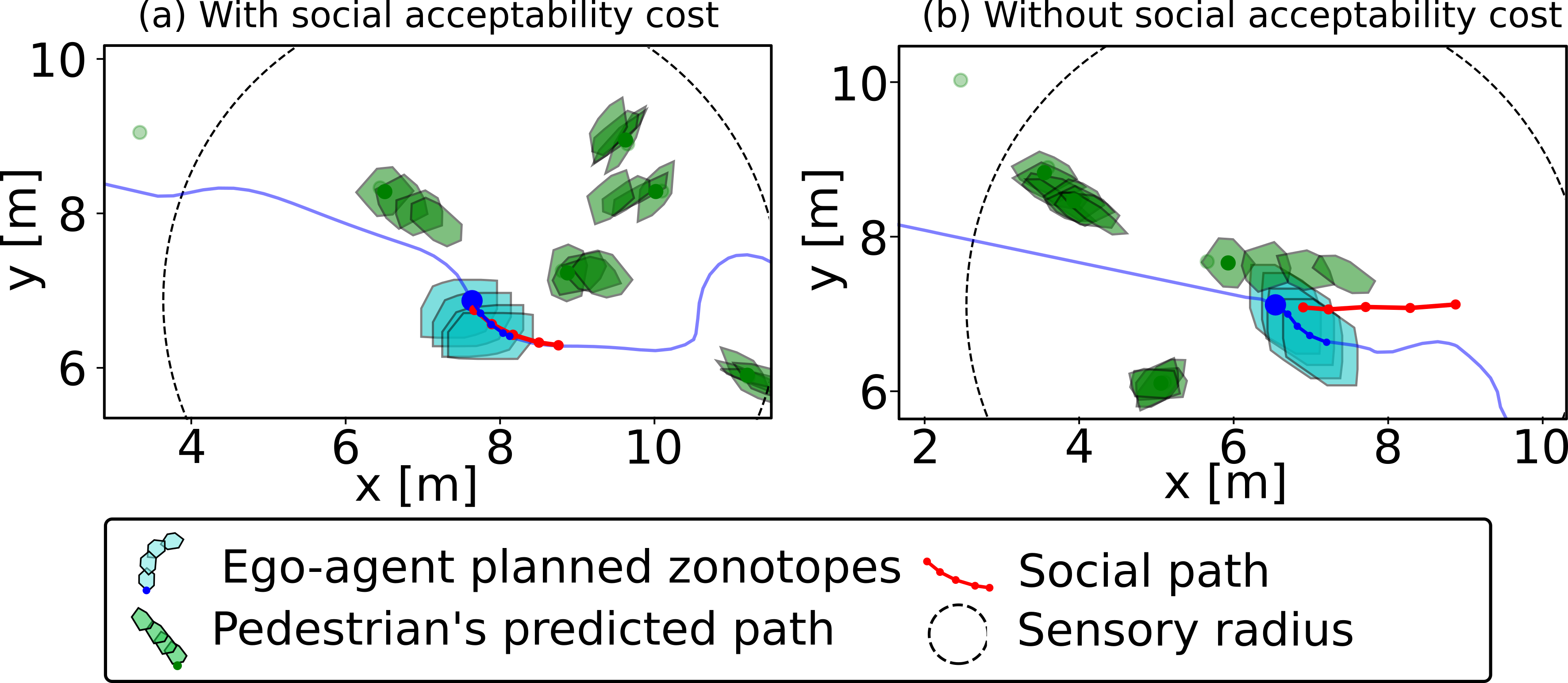}}
\caption{Qualitative results of social acceptability: comparison of different social acceptability levels. (a) shows a socially acceptable trajectory generated by SZN-MPC with social acceptability cost~(\ref{eq:socia_cost}), as the ego-agent's planned path (shown in blue) follows the predicted social path of ESN (shown in red), while (b) shows a trajectory generated by SZN-MPC without social acceptability cost~\eqref{eq:socia_cost} where a larger deviation is observed between the ego-agent's planned path and the predicted social path of ESN.}
\label{fig:qual_social_metric}
\end{figure}

To quantify social acceptability we rely on displacement errors between the ground truth data $\boldsymbol{c}^{\rm ego}_i$ and ESN prediction output $\mathcal{T}^{\rm ego}_{{\rm mid},i}$.
\begin{equation}
    {\rm ADE} = \frac{\sum_{i=1}^{t_{f}-1} \| \mathcal{T}^{\rm ego}_{{\rm mid},i} - \boldsymbol{c}^{\rm ego}_i \|}{t_{f}-1}
\end{equation}
Fig.~\ref{fig:quan_social_metric}(a) shows that ESN produces an ADE$=0.218$ m over the prediction horizon of $7$ timesteps\footnote{The prediction horizon timesteps is $7$ and not $8$, since the displacement error is calculated based on the middle points $\mathcal{T}^{\rm ego}_{\rm mid}$ of $\mathcal{T}^{\rm ego}_{[1, 8]}$}, and FDE$= 0.447$ m.

For Digit to achieve a socially acceptable path, it needs to track the path produced by ESN, by assuming that ESN's ADE $= 0.218 < \epsilon$ as defined in Definition.~\ref{def:social_path}.
Fig.~\ref{fig:quan_social_metric}(b) shows that we are able to reduce the social acceptability cost when running SZN-MPC~\eqref{eq:problem} with the social running cost Eq.~(\ref{eq:socia_cost}), thus, promoting social acceptability by tracking the path of ESN. Fig.~\ref{fig:qual_social_metric} shows the results of a planned trajectory of SZN-MPC with and without social acceptability cost~\eqref{eq:socia_cost}.

\subsubsection{Locomotion Feasibility}

\begin{figure}[t]
\centerline{\includegraphics[width=.95\columnwidth]{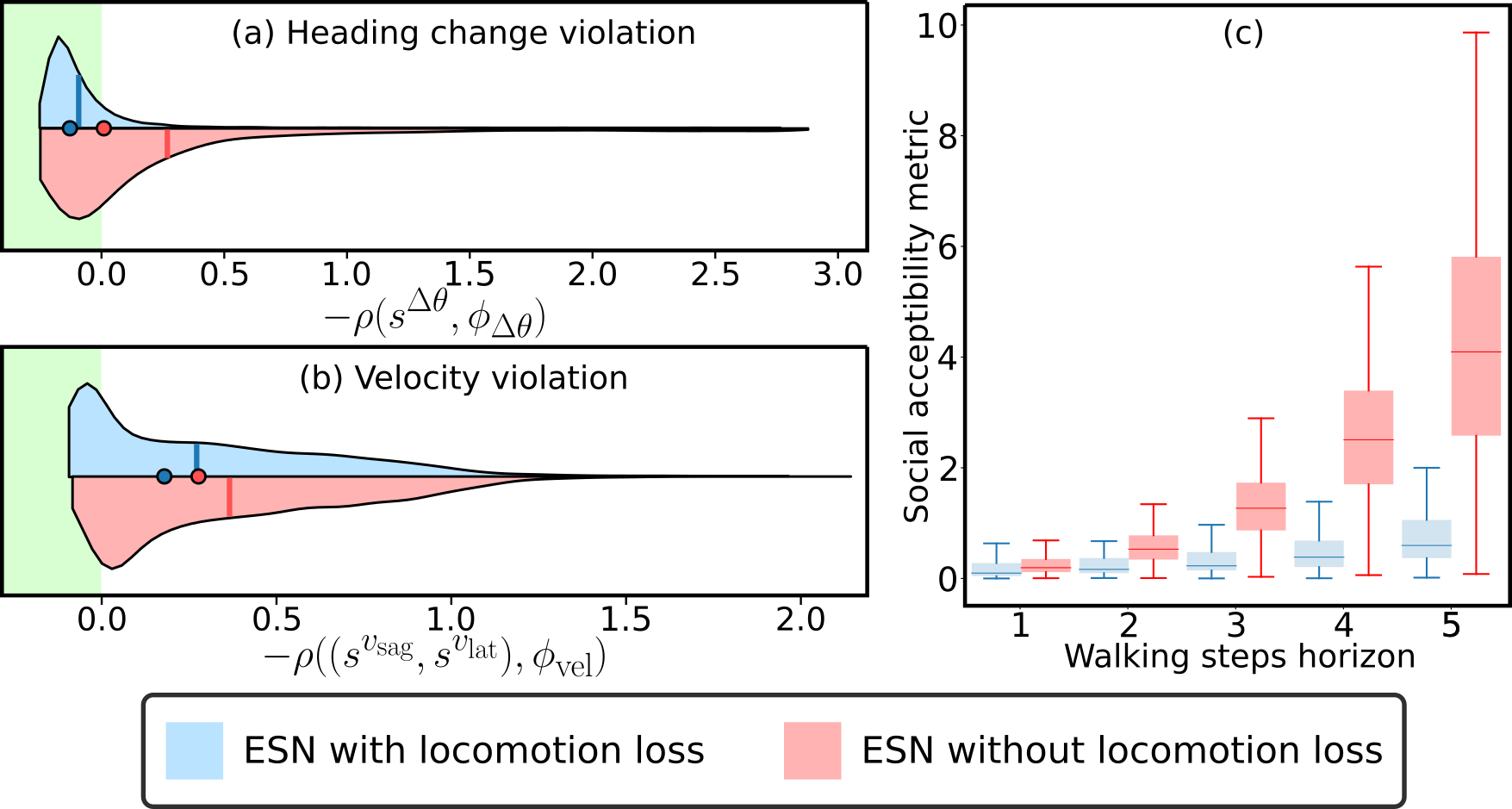}}
\caption{Violin plots of ESN test results with and without STL locomotion losses are shown in blue and red, respectively. The shaded green region is where the heading change specification (a), and locomotion velocity specification (b) are satisfied, i.e., $\rho(s^{\Delta \theta},\phi_{\Delta \theta}))<0$ and $-\rho((s^{v_{\rm sag}},s^{v_{\rm lat}}),\phi_{\rm vel}) < 0$. The solid vertical line represents the mean value of $-\rho$, while the dot represents the median. The data is collected based on the UNIV dataset with $7831$ unique frames. (c) Shows the social acceptability metric when ESN is integrated with SZN-MPC with and without the locomotion losses.}
\label{fig:stl_loss}
\end{figure}

Integrating the locomotion losses into ESN training limits the violation of the locomotion safety specifications~(\ref{eq:velocity}) and~(\ref{eq:heading_change}) as seen in Fig.~\ref{fig:stl_loss}(a)-(b).
The reduction in the violation allows SZN-MPC to generate a trajectory that achieves an improved tracking of the socially acceptable path as shown in Fig.~\ref{fig:stl_loss}(c). 
These results indicate that the locomotion losses shape ESN output to comply with the capabilities of bipedal locomotion.

\subsection{Experiment 2: Benchmarking}

\begin{figure}[t]
    \centering    \includegraphics[width=0.8\linewidth]{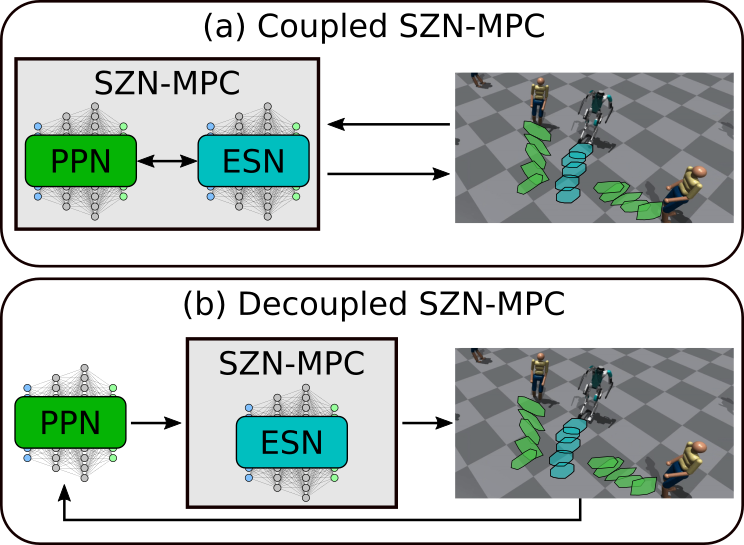}
    \caption{Block diagram of (a) coupled SZN-MPC and (b) decoupled SZN-MPC.}
    \label{fig:mpc_block}
\end{figure}

\begin{figure*}[t]
\centerline{\includegraphics[width=.95\textwidth]{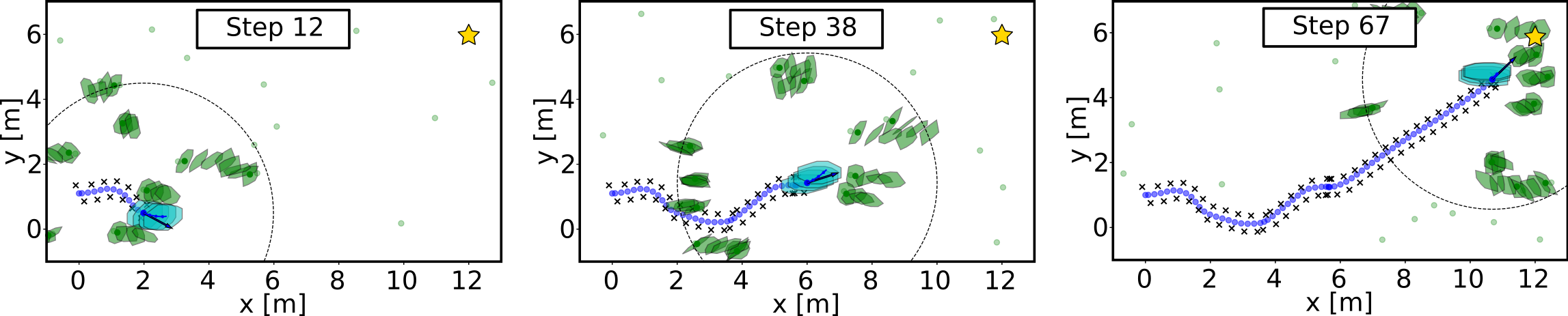}}
\caption{Storyboard snapshots of SZN-MPC trajectory at different walking steps. The ego-agent (cyan) successfully reaches the goal (yellow $\star$) while avoiding pedestrians (green). The dashed circle is the sensory radius of the ego-agent. Blue dots represent the CoM of ROM, and black $\times$ is the desired foot placement.
Green dots are unobserved pedestrians.}
\label{fig:frames}
\end{figure*}

\begin{figure*}[t]
\centerline{\includegraphics[width=.98\textwidth]{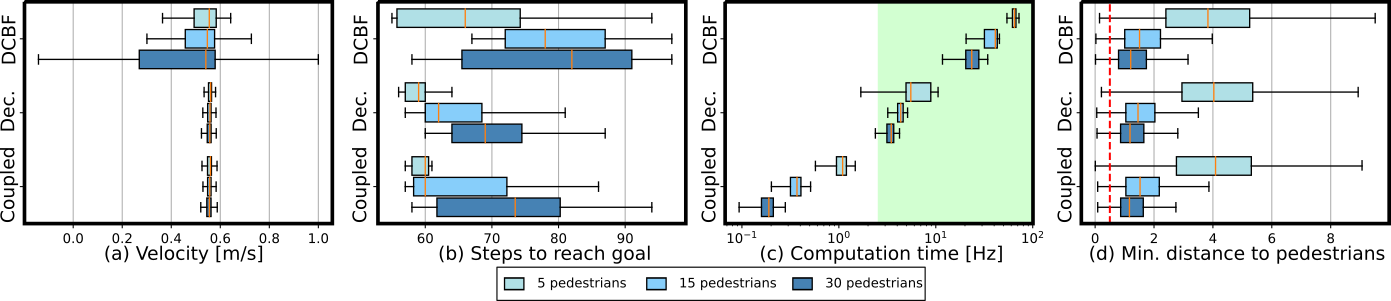}}
\caption{Benchmarking results for the ego-agent's (a) velocity, (b) optimality: number of walking steps to reach within $1$ m of the final goal, (c) frequency, and (d) safety: minimum distance to pedestrians. The data consists of 20 different trials with random initial conditions and a fixed goal location (see Sec.~\ref{subsubsec:test_env}). Each trial is limited up to $100$ walking steps. The velocity data is collected before reaching the goal, to avoid collecting a stopping velocity. The frequency is calculated based on a data collection of $300$ walking steps, and the green shaded area in (c) is the minimum required computation time of SZN-MPC for Digit hardware implementation. (d) the dashed red line is $0.5$ m which is $r$ in (\ref{eq:cbf_h}), and what is considered as $d$ in navigation safety in Definition.~\ref{def:nav_safety}. }
\label{fig:benchmark}
\end{figure*}

We compare a coupled and decoupled version of our method with a baseline approach for LIP-based MPC with a discrete-time control barrier function (DCBF-MPC) for collision avoidance of dynamic obstacles~\cite{narkhede2022sequential, teng2021toward}. We compare the conservativeness of the produced trajectory, social acceptability, safety and optimality, and finally the computational cost. Snapshots of SZN-MPC results at different walking steps are shown in Fig.~\ref{fig:frames}.

\subsubsection{Models}
\paragraph{Coupled SZN-MPC}
In this setup, the prediction (PPN) and planning (ESN) networks are coupled during the trajectory optimization, where the MPC reasons about the effect of the ego-agent's control on the future prediction of the pedestrians.
A block diagram of this model is shown in Fig.~\ref{fig:mpc_block}(a).

\paragraph{Decoupled SZN-MPC (dec.)}
In this setup, PPN and ESN are decoupled. Before each solve of (\ref{eq:problem}), the pedestrians' future prediction is conditioned on the optimal solution from the previous MPC solve for the ego-agent and is fixed throughout the optimization for the current solve. Namely, the PPN module is queried only once for each MPC solve.
A block diagram of this model is shown in Fig.~\ref{fig:mpc_block}(b).

\begin{rem}
    For clarification, our decoupled SZN-MPC is not considered as the category of the decoupled social navigation literature as explained in Sec.~\ref{RW:social_nav}, since the PPN module is conditioned on the ego-agent next planned walking step. Thus, our decoupled SZN-MPC can be viewed as a coupled social navigation method in the literature.
\end{rem}

\paragraph{DCBF-MPC}
We compare our path plan to that generated by a LIP-based MPC with a DCBF for navigation safety, where we substitute ~\eqref{eq:HS_mpc_reach} and ~\eqref{eq:HS_mpc_avoid} with a DCBF:
\begin{equation}
     h(\boldsymbol{p}^{\rm ego}_{q+1},\boldsymbol{p}_{k_{q+1}}) \geq (1-\gamma)  h(\boldsymbol{p}^{\rm ego}_{q},\boldsymbol{p}_{k_{q}}), \; \forall \; k_q
\end{equation}

where $(1-\gamma)$ is a class $\mathcal{K}_{\infty,e}$, $0 < \gamma \leq 1$ \cite{agrawal2017discrete,zeng2021safety}, and $h(\boldsymbol{p}^{\rm ego}_{q},\boldsymbol{p}_{k_{q}})$ is a distance metric defined as:

\begin{equation}
\label{eq:cbf_h}
   h(\boldsymbol{p}^{\rm ego}_q,\boldsymbol{p}_{k_{q}})=  \|\frac{1}{r}( \boldsymbol{p}^{\rm ego}_q-\boldsymbol{p}_{k_{q}}) \|  -1
\end{equation}
and the optimization is initialized with $h(\boldsymbol{p}^{\rm ego}_{0},\boldsymbol{p}_{k_{0}}) \geq 0 $. 
For this model, we also use PPN as a prediction module for the surrounding pedestrians, and SGAN as the pedestrian simulator. 

\begin{figure*}[t]
\centerline{\includegraphics[width=.9\textwidth]{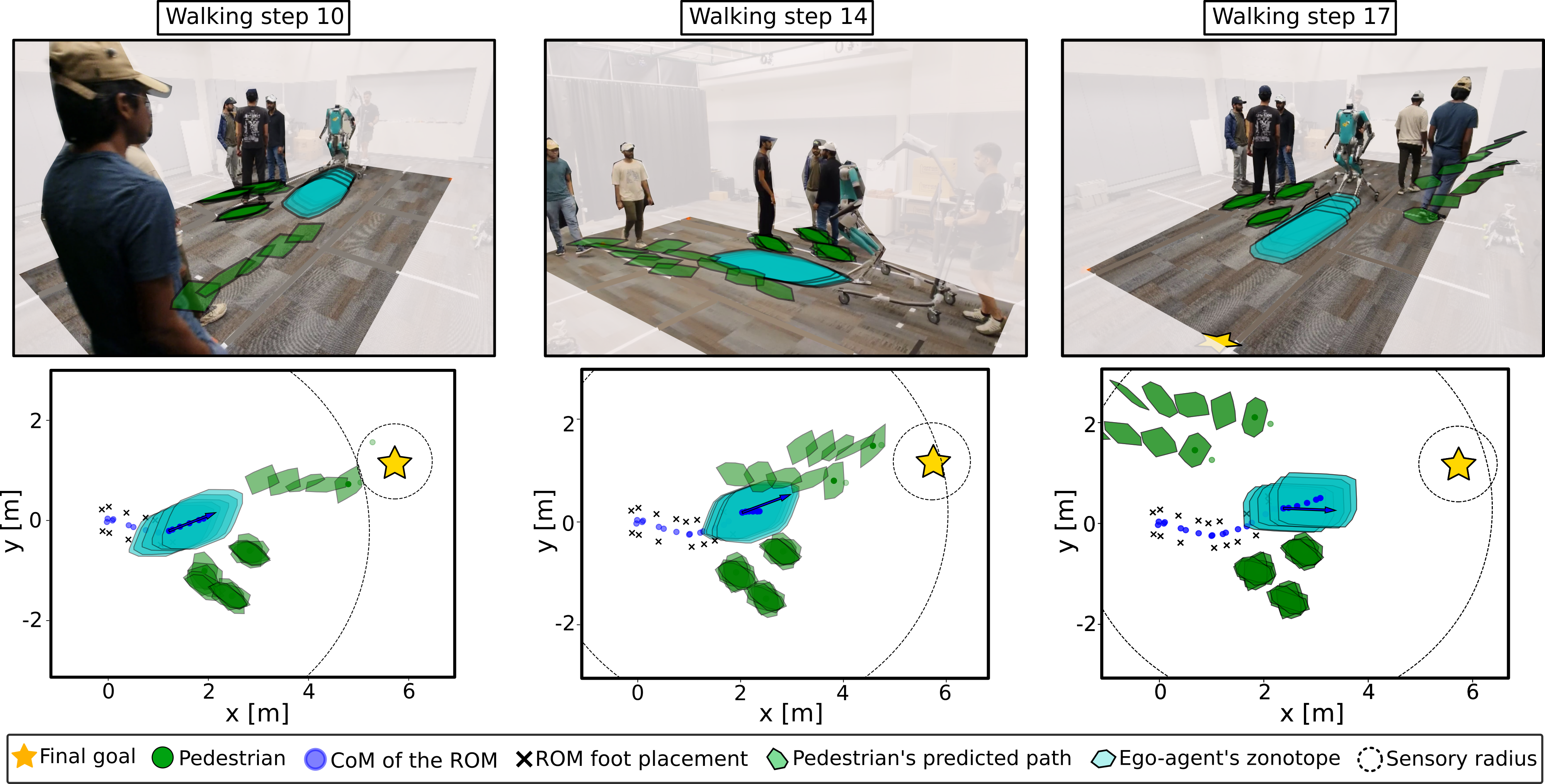}}
\caption{A storyboard snapshots of the hardware experiments with $5$ pedestrians. In this experiment, the first group consists of three stationary pedestrians standing in place, while the other group consists of two pedestrians walking towards Digit.}
\label{fig:hardware_story}
\end{figure*}
\subsubsection{Conservativeness}

In Fig.~\ref{fig:benchmark}(a), all three models produce relatively similar median velocities over the testing data.
However, coupled SZN-MPC and decoupled SZN-MPC produce more consistent velocities.
Our proposed method is more efficient as shown by the number of steps taken to reach the goal Fig.~\ref{fig:benchmark}(b), with
less variation in velocity and similar safety performance (see Fig.~\ref{fig:benchmark}(a) and (d)).
The high variability in velocity that DCBF-MPC produces indicates adaptability in a dynamic environment, however, it does not translate to a safer path in a densely crowded environment (instead, it generates the same level of safety as SZN-MPC) as it produces a similar minimum distance to the surrounding pedestrians (See Fig.~\ref{fig:benchmark}(a) and (d)).

\subsubsection{Social Acceptability} 
SZN-MPC produces a more consistent and predictable behavior for the ego-agent when compared to DCBF-MPC as indicated by the the smaller interquartile range in Fig.~\ref{fig:benchmark}(a).
Predictability of the ego-agent behavior in a social context is desirable by pedestrians as it is perceived to be less disruptive.
\begin{rem}
    The coupled SZN-MPC and decoupled SZN-MPC produce similar results. The reason is as follows: Although the decoupled SZN-MPC only inquires the PPN module once for each MPC solve and then fixes the pedestrian prediction during the optimization, this MPC is solved at a frequency of individual walking steps and each solve will couple the pedestrian prediction with the ego-agent planning. Thus, the decoupled SZN-MPC presents a similar level of coupling between the pedestrian prediction and ego-agent planning, and produces comparable results to the coupled SZN-MPC.
\end{rem}

\subsubsection{Safety and Optimality}
All three methods produce comparable safety performance by maintaining a similar minimum distance to the pedestrians as shown in Fig.~\ref{fig:benchmark}(d).
However, SZN-MPC consistently generates a more optimal path, as indicated by the lower number of steps taken to reach the goal (see Fig.~\ref{fig:benchmark}(b)).
Thus, we are not sacrificing safety by moving faster.
SZN-MPC generates optimal paths for the ego-agent due to its proactive approach, i.e., the SZN-MPC generates future reachable sets that take into account not only the collision avoidance but also social influence from all surrounding pedestrians. On the other hand, DCBF-MPC is reactive, as it only reacts to the PPN predictions of the pedestrians to avoid collisions. This reactivity leads to the ``freezing robot problem"~\cite{trautman2015robot}, as evidenced by the low velocities that DCBF-MPC generates, specifically in environments with high crowd densities~\ref{fig:benchmark}(a).
\begin{rem}
    The pedestrian simulation lacks assurance of generating non-adversarial paths.
    Despite considering the ego-agent's position, there are occasional cases where the simulator generates paths for pedestrians that intersect with the ego-agent and other pedestrians.
    In our real-world experiments, we operate under the assumption of pedestrian collaboration and non-adversarial behavior towards the ego-agent, following social norms.
\end{rem}

\subsubsection{Computational Cost}
The three models' computational costs are different by orders of magnitude, where DCBF-MPC is $10^2$ times faster than the coupled SZN-MPC and $10$ times faster than the decoupled SZN-MPC as shown in Fig.~\ref{fig:benchmark}(c).
Note that, the MPC is used as a step planner for Digit, and it only needs to be solved once before the swing phase of the walking motion ends. Since the commanded step duration is $0.4$ s, which is the minimum computational time required for hardware implementation, the SZN-MPC needs to be solved at a rate of at least $2.5$ Hz  (Sec.\ref{sec:lowlevel_cont}).
Based on the testing parameters (Sec.\ref{sec:implementation}), the coupled SZN-MPC is not viable for hardware implementation. Therefore, we conclude the decoupled SZN-MPC outperforms the coupled version, and will use the decoupled SZN-MPC for our hardware implementation.
Computational efficiency can be increased with a smaller number of the planning horizon $N$, and the number of pedestrians in the environment.

%% file: sections/11_hardware.tex
\begin{figure*}[t]
\centerline{\includegraphics[width=.9\textwidth]{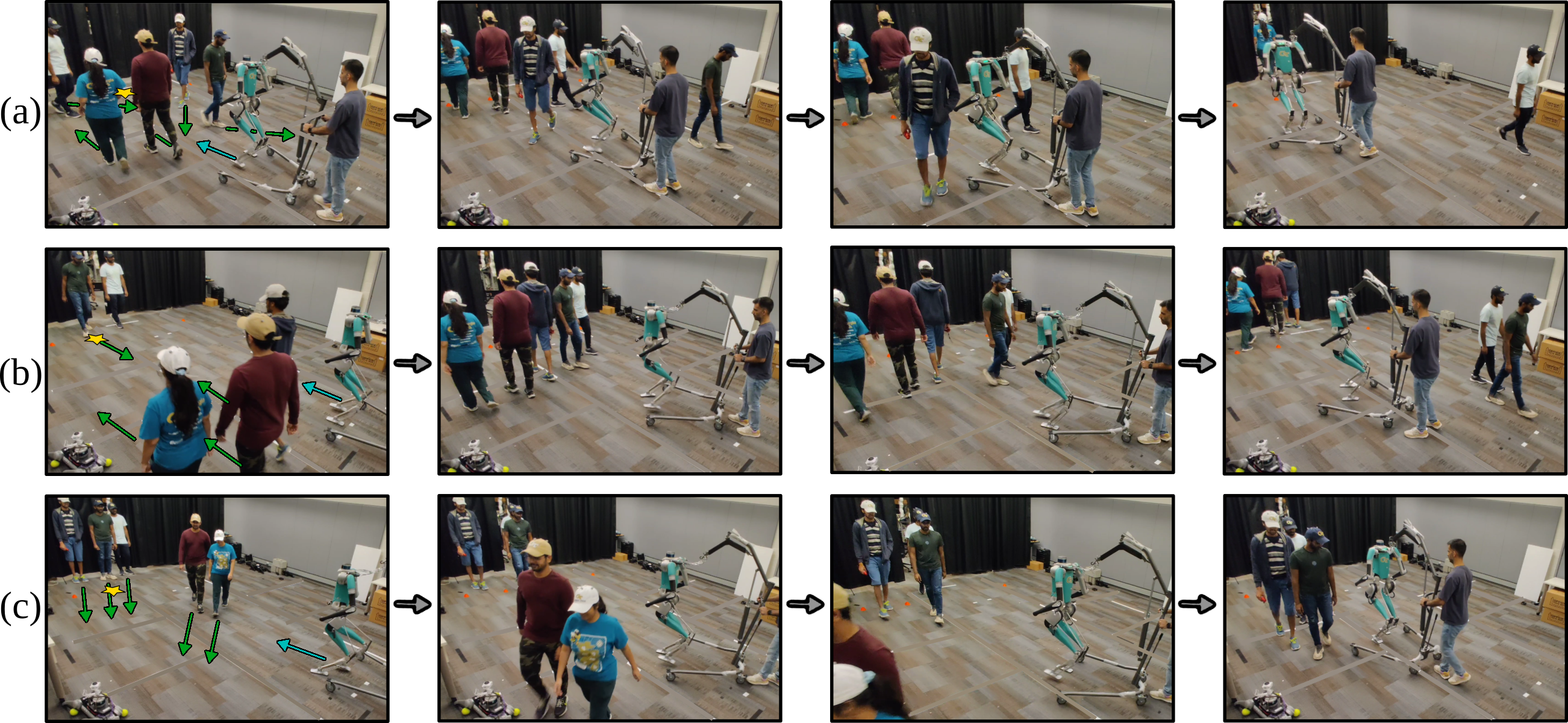}}
\caption{Snapshots of three different testing scenarios. (a) depicts pedestrians walking in different directions. (b) illustrates a group of pedestrians walking alongside Digit in a row while the other group walks towards Digit. (c) demonstrates two groups walking in the direction opposite to Digit. In the first column, green arrows indicate the direction of the pedestrians' movement, while a cyan arrow indicates Digit's direction, and the yellow star is the final goal position for Digit.}
\label{fig:hardware_scenarios}
\end{figure*}

\subsection{Experiment 3: Hardware Validation in A Human Crowd}
We conduct a series of hardware experiments to achieve the following objectives:
First, to showcase the versatility of our proposed method in reliably navigating social settings with pedestrians across various scenarios. 
Second, to validate our network architecture choices, specifically learning the pedestrians' future path prediction and the ego-agent's future path directly as zonotopes, which facilitates real-time implementation of the full framework on Digit's onboard PC.

\begin{rem}
    The personal space modulation and modeling error compensation in Sec.~\ref{sec:zonotope_refinement}, are included in this hardware experiment for their practical use. Without such refinements, Digit's trajectory was relatively close to the surrounding pedestrians. A key feature of the zonotope parametrization is that it allows us to compensate for desired behaviors downstream without the need to retrain the SZN.
\end{rem}

\subsubsection{Low-level Full-Body Control}\label{sec:lowlevel_cont}
At the low level we use the Angular momentum LIP planner introduced in~\cite{Gong2022AngularMomentum}, and a Digit's passivity controller with ankle actuation which we have previously shown to exhibit desirable ROM tracking results~\cite{shamsah2023integrated}.
Here we set the desired walking step time and the desired lateral step width to be $0.4$ s and $0.4$ m, respectively.

\subsubsection{Experiment Setup}

Our experimental setup uses an indoor VICON motion capture system~\cite{Vicon} to measure the current position of the surrounding pedestrians $\position_{k_0}$.
The coordinates of pedestrians are sent over to SZN-MPC every $0.4$ s to match the human crowd dataset SZN is trained on. SZN-MPC then solves for the ego-agent's social trajectory and sends the next desired CoM velocity and heading change $(v\local\nextq, \theta\nextq)$ to the low-level controller.
The prediction, planning, and low-level control are executed on Digit's onboard PC and in real-time, with a walking step horizon $N= 4$.
The experimental space in $6 \times 3$ m$^2$, Digit's starting position is $\position\ego_0 = (0, 0)$ m with $\theta_0 = 0^\circ$, and the goal location is diagonally across the experiment space at  $\mathcal{G}=(5.72, 1.18)$ m (see Fig.~\ref{fig:hardware_story}).

\subsubsection{Experiments}
Our framework is demonstrated in various mock social scenarios\footnote{videos of the experiments are found here~\href{https://youtu.be/bUSbsj3cCH4?si=wPDV0ezDONasbZNm}{\nolinkurl{https://youtu.be/bUSbsj3cCH4?si=wPDV0ezDONasbZNm}}. 
}.
In Fig.~\ref{fig:hardware_story}, we depict a social scenario involving two groups of pedestrians.
The first group consists of three stationary pedestrians, while the other group consists of two pedestrians walking towards Digit.
Our results show that SZN accurately predicts the two groups' behaviors and safely navigates toward the goal location marked by the yellow star. Digit adjusts its position closer to the stationary group to avoid collision with the approaching pedestrians.
Three different scenarios are also shown in Fig.~\ref{fig:hardware_scenarios}.
In Fig.~\ref{fig:hardware_scenarios}(a), pedestrians move towards the center of the space from different directions and continue their path out of the experimental space. In Fig.~\ref{fig:hardware_scenarios}(b), one group walks alongside Digit in a row while the other group of pedestrians walks in the opposite direction.
In Fig.~\ref{fig:hardware_scenarios}(c), two groups of pedestrians walk towards Digit but with a slightly different lateral direction.
In all scenarios, Digit successfully generate socially acceptable and safe navigation paths while maintaining a proactive forward motion.

\begin{figure}[!t]
\centerline{\includegraphics[width=.99\linewidth]{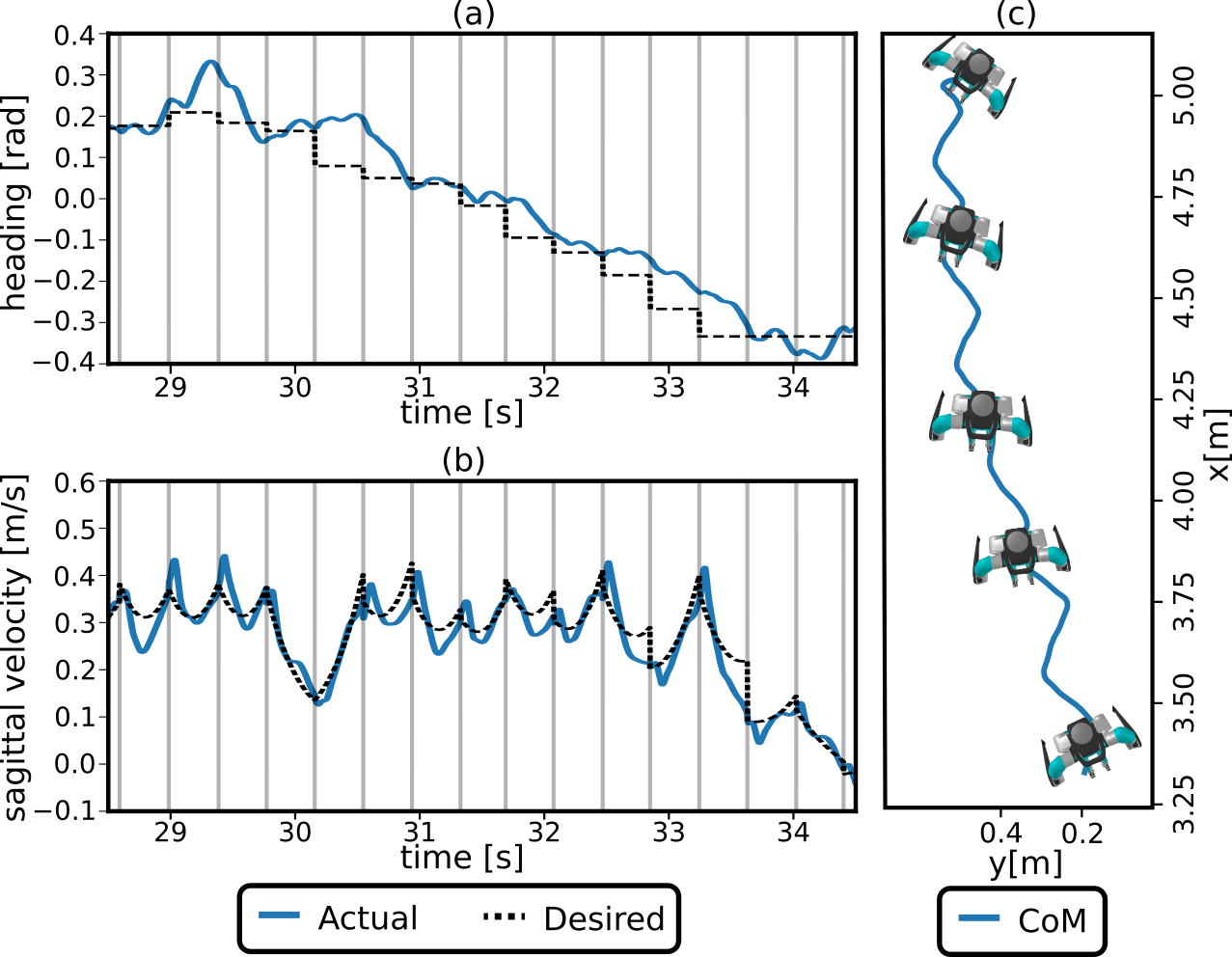}}
\caption{Hardware experiment results: in (a) we show the torso heading tracking performance, and CoM sagittal velocity tracking performance in (b), and in (c) we show the experimental CoM position with superimposed Digit illustration. The vertical gray lines indicate the foot contact switching instants of each walking step.}
\label{fig:hardware_velocity}
\end{figure}

Fig.~\ref{fig:hardware_velocity} illustrates the tracking performance comparisons between the desired $(v\local, \theta)$ from SZN-MPC and the actual hardware responses of Digit.
Fig.~\ref{fig:hardware_velocity}(a) shows the tracking performance of the heading for Digit $\theta$, and Fig.~\ref{fig:hardware_velocity}(b) shows the CoM sagittal velocity $v\local$ tracking performance.
The continuous desired velocity profile is the desired continuous ROM velocity based on $v\local$~\cite{shamsah2023integrated,Gong2022AngularMomentum}.
SZN-MPC updates the target parameters $(v\local, \theta)$ at every walking step, with vertical gray lines marking the foot contact events.
Discrepancies are bound to occur when using a ROM plan to control a full-order system.
These issues become especially noticeable with elements like body orientation (see Fig.~\ref{fig:hardware_velocity}(a)), which the ROM does not explicitly account for.

The hardware experiments demonstrate the efficacy of our proposed approach. The design choices made for the SZN architecture, the ROM-based motion planner, and the reachable-based collision avoidance collectively enable safe and real-time experimentation with the hardware.

%% file: sections/12_limitations.tex
\section{Limitations and Discussions}\label{sec:discussion}
In this section, we discuss the limitations of our proposed framework, including the assumptions underpinning our social acceptability evaluation, the computational demands of our approach, and the constraints related to the locomotion gait of our bipedal robot.

\subsection{Social Acceptability Evaluation and Datasets}
The core assumption underlying our framework is that the paths recorded in human crowd datasets ~\cite{lerner2007crowds, pellegrini2009walk} represent socially acceptable trajectories.
Although this assumption may hold true, we acknowledge that in practical, real-world implementations, human crowds may behave differently in the presence of a bipedal robot.
This presents a significant challenge. Collecting data on human's social reaction with Digit does not ensure that pedestrians will behave naturally.
Since the presence of bipedal robots in social settings is still relatively new and potentially intimidating, it may influence pedestrian's social navigation behavior.
One way to potentially address this issue is using a human pedestrian simulation in which the simulated pedestrians perceive the robot as just another pedestrian within the scene, similar to the approach used in this work.
However, accurately simulating pedestrian interactions remains a challenging task.

We base the design of socially acceptable paths for the ego-agent on the past paths of pedestrians.
To achieve a more comprehensive socially acceptable behavior for the ego-agent, further considerations could include (1) the design of a natural walking gait for the robot such that the pedestrians are more at ease around the robot ego-agent; (2) an analysis of the facial expressions of surrounding pedestrians and adjusting the ego-agent's path accordingly; and (3) incorporation of the social scenario context such as shopping mall, airport, and hospital.
\subsection{Computational Cost and Outdoor Implementation}
We opted for a relatively straightforward neural network architecture to enable real-time implementation. However, processing large crowds in real-time can be challenging.
This issue could be mitigated by employing a more advanced filtering scheme (rather than the simple circle threshold used in our study) to exclude pedestrians that do not impact the ego-agent’s path. Ideally, this new filtering scheme could enable an optimized neural network design that does not increase the computational burden during the trajectory optimization phase.
Additionally, outdoor deployments may increase computational demands due to the complexity of the environment, which includes not only pedestrians but also static and dynamic obstacles and other environmental components.
Using sensors to track pedestrians and processing this data on the fly for the SZN-MPC could extend the solve time beyond acceptable limits.
\subsection{Locomotion Gait}
In our implementation, we constrain the walking gait to maintain stable torso Euler angles and fixed arm motion.
In human locomotion, particularly in narrow spaces such as corridors, socially acceptable behaviors often involve adjusting the torso yaw angle while continuing straight center-of-mass motion.
By adjusting the torso, a person can reduce their effective width, making it easier to pass by others without direct contact. This subtle change in body orientation helps to communicate intent and awareness of social norms, facilitating smoother and more courteous interactions in confined areas. In scenarios where forward movement is obstructed by standing pedestrians, the capability to walk sideways could provide a more efficient and socially acceptable solution to navigate around other pedestrians. Without this capability, the robot is forced to either come to a complete stop or walk backward.
\subsection{Environment and Task}
Our framework only considers an open space environment with no static obstacles and a simple reach-avoid task.
To expand this capability, more complex navigation tasks in an environment with both dynamic and static obstacles can be implemented by using high-level planners hierarchically connected as a layer on top of the SZN-MPC, e.g., formal task planning methods as in~\cite{shamsah2023integrated}.
This hierarchical approach allows for better handling of more sophisticated tasks and possibly adversarial environmental events.
Furthermore, the modeling error GP in Sec.~\ref{subsec:Modelling_error_refinements} can be expanded to include terrain profile uncertainty, enabling the robot to navigate complex and rough terrain~\cite{jiang2023abstraction, muenprasitivej2024bipedal}.

%% file: sections/13_conclusion.tex
\section{Conclusion}\label{sec:conclusion}
Integration of bipedal robotic systems in real-world environments is still an open problem, specifically in human-oriented environments. 
These environments are less predictable and require a nuanced level of social interaction between the robot and humans.

We present a novel approach for socially acceptable bipedal navigation.
At the core of our framework is SZN, a learning architecture for reachability-based prediction of future pedestrian reachable sets and planning a socially acceptable reachable set for the robot parameterized as zonotopes. Zonotopes allow for efficient modulation of the reachable set based on modeling uncertainty and personal space preferences.
Our integration of SZN into MPC allows for real-time pedestrian trajectory prediction with bipedal motion planning.
SZN-MPC optimizes over the output of the neural network, with a novel cost function designed to encourage the generation of socially acceptable trajectories, striking a balance between efficient navigation and adherence to social norms.
The extensive validation through simulations and hardware experiments solidifies the effectiveness of the proposed framework.

%% file: appendix.tex
\section{Derivation of Linear Inverted Pendulum model }
\label{appendix:LIP}
In a manner similar to the derivation of the step-to-step discrete Linear Inverted Pendulum (LIP) dynamics as in~\cite{teng2021toward, narkhede2022sequential}, the continuous motion of the LIP dynamics in the sagittal direction is governed by $\ddot{x}\local=-(g/H)u^f$, where $g$ is the gravitational acceleration, $H$ is CoM height, and $u^f$ is the sagittal distance of the stance foot from CoM.
The closed-form solution is given by
\begin{equation*}
    \begin{bmatrix}
    x\local(t) \\
    v\local(t)
    \end{bmatrix} 
    =
    \begin{bmatrix}
        1 & \sinh(\omega t)/\omega \\
        0 & \cosh(\omega t) 
    \end{bmatrix}
    \begin{bmatrix}
        x\local(0) \\
        v\local(0)
    \end{bmatrix} +
    \begin{bmatrix}
        1 -\cosh(\omega t) \\
        -\omega \sinh(\omega t)
    \end{bmatrix} u^f
\end{equation*}
where $\omega = \sqrt{g/H}$.
Setting each step duration to a constant $T$, such that the state at the $(q+1)^{\rm th}$ step is $x\nextq = x\currq(T)$, we obtain the step-to-step discrete LIP model as:
\begin{equation*}
    \begin{bmatrix}
    x\local\nextq \\
    v\local\nextq
    \end{bmatrix} 
    =
    \begin{bmatrix}
        1 & \sinh(\omega T)/\omega \\
        0 & \cosh(\omega T) 
    \end{bmatrix}
    \begin{bmatrix}
        x\local\currq \\
        v\local\currq
    \end{bmatrix} +
    \begin{bmatrix}
        1 -\cosh(\omega T) \\
        -\omega \sinh(\omega T)
    \end{bmatrix} u^f\currq
\end{equation*}

Therefore the sagittal CoM position change in one walking step can be expressed as:
\begin{equation*}
    \Delta x\local= x\local\nextq - x\local\currq = v\local \sinh(\omega T)/\omega  +(1 -\cosh(\omega T))u^f\currq
\end{equation*}

To obtain the dynamics $\state = (\position, v\local, \theta)$, where $\position = (x,y)$, we introduce the heading change based on $u^{\Delta \theta} \currq = \theta\nextq - \theta_{q}$ to the sagittal dynamics and obtain the following set of dynamics as in~(\ref{eq:lip_dynamics}):
\begin{subequations}
\begin{align*}
x\nextq &= x_q +\Delta x\local\cos(\theta_q) \\
y\nextq &= y_q + \Delta x\local\sin(\theta_q) \\
v\local\nextq &= \cosh(\omega T) v\local\currq - \omega \sinh(\omega T) u^f\currq \\ 
\theta\nextq &= \theta\currq + u^{\Delta \theta}\currq 
\end{align*}
\end{subequations}

\section{Network Architecture Parameters}
\label{app:network_arch}
\begin{table}[!h]
    \centering
    \caption{Network architecture parameters}
    \begin{tabular}{|c|c|}
       \hline
        \multicolumn{2}{|c|}{Pedestrian Prediction Network} \\
        \hline
       $E_{\rm ped}$ & $16\rightarrow32\rightarrow16$  \\
       $E_{\rm end}$ & $2 \rightarrow 8 \rightarrow 16$  \\
       $E_{\rm nxt}$ & $2 \rightarrow 32 \rightarrow 16$  \\
       $P_{\rm future}$  & $50 \rightarrow 32 \rightarrow 16 \rightarrow 32 \rightarrow 70$ \\
        $E_{\rm latent}$ & $48 \rightarrow 8 \rightarrow 16 \rightarrow 32$  \\
        $D_{\rm latent}$ & $48 \rightarrow 32 \rightarrow 16 \rightarrow 32 \rightarrow 2$ \\
         \hline
         \hline
        \multicolumn{2}{|c|}{Ego-agent Social Network} \\
        \hline
       $E_{\rm goal}$ & $2\rightarrow 8 \rightarrow 16 \rightarrow2$  \\
       $E_{\rm future}$ & $16 \rightarrow 64 \rightarrow 32 \rightarrow 16$  \\
       $E_{\rm nxt}$ & $2 \rightarrow 64 \rightarrow 32 \rightarrow 2$  \\
       $E_{\rm traj}$  & $16 \rightarrow 64 \rightarrow 32 \rightarrow 16$ \\
        $E_{\rm latent}$ & $36 \rightarrow 8 \rightarrow 50 \rightarrow 16$  \\
        $D_{\rm latent}$ & $36 \rightarrow 128 \rightarrow 64 \rightarrow 128 \rightarrow 70$ \\
         \hline
    \end{tabular}

    \label{tab:network_arch}
\end{table}

%% file: Acknowledgment.tex
\section*{Acknowledgment}
The authors would like to express their gratitude to Dr. Seth Hutchinson and Dr. Sehoon Ha for allowing us to use their lab space and motion capture system. Special thanks to Nitish Sontakke for helping with the motion capture system setup and to Vignesh Subramanian, Abivishaq Balasubramanian, Meghna Narayanan, Hemanth Sai Surya Kumar Tammana, Harini Mudradi, Venkat Sai Chinta, Manthan Joshi for their participation in the hardware experiments. 